\documentclass[letterpaper,10pt,journal,twocolumn]{IEEEtran}
\usepackage{amsmath}
\usepackage{amsfonts}
\usepackage{amssymb}
\usepackage{adjustbox}
\usepackage{booktabs}
\usepackage{multirow}
\usepackage{graphicx}
\usepackage{hyperref}
\usepackage{multicol}
\usepackage{subcaption}
\usepackage{xcolor}

\usepackage[linesnumbered,ruled,vlined]{algorithm2e}

\newcommand{\algname}{EXACT-MPPI}
\newcommand{\vect}[1]{\boldsymbol{\mathbf{#1}}}

\begin{document}

\title{EXACT-MPPI: Exact Signed-Distance Navigation for Arbitrary-Footprint Robots from Point Clouds via Path Integral Control}

\author{Chen~Peng$^{\dagger}$,
    Zhikang~Ge$^{\dagger}$,
    Wenwu~Lu,
    Haiming~Gao$^{\ast}$,
    Stavros~Vougioukas,
    and~Peng~Wei$^{\ast}$
    \thanks{$^{\dagger}$Equal contribution. $^{\ast}$Corresponding author.(e-mail: ghm@zju.edu.cn; penwei@ucdavis.edu)}%
    \thanks{Chen Peng is with the ZJU-Hangzhou Global Scientific and Technological Innovation Center, Zhejiang University, Hangzhou, China, and also with the College of Biosystems Engineering and Food Science, Zhejiang University, Hangzhou, China.}%
    \thanks{Zhikang Ge, Wenwu Lu, and Haiming Gao are with the ZJU-Hangzhou Global Scientific and Technological Innovation Center, Zhejiang University, Hangzhou, China.}%
    \thanks{Stavros Vougioukas and Peng Wei are with the Department of Biological and Agricultural Engineering, University of California, Davis, Davis, California, USA.}%
    \thanks{Project website: \url{https://agroboticsresearch.github.io/exact-mppi/}. Code: \url{https://github.com/caseypen/EXACT-mppi}.}%
}

\maketitle

\begin{abstract}

    Ground robots often carry payloads, implements, or other attachments that turn their effective footprint into complex, non-convex shapes. Navigating safely through clutter then requires reasoning about this true geometry, yet most local planners simplify it with convex or inflated proxies and rasterize sensor data into occupancy grids or distance fields. Both choices eliminate feasible motions when clearance is comparable to the footprint geometry. We present \algname{}, a training-free local navigation framework that maps local point-cloud observations and sparse guidance directly to motion commands, without any intermediate map representation. The framework embeds an analytic, exact signed-distance evaluator into a Model Predictive Path Integral (MPPI) controller. The footprint is represented as a simple polygon for general convex or concave planar shapes, with a rectangle-cover specialization for faster evaluation of rectilinear footprints, enabling footprint-aware collision costs without convex decomposition, inflation, or learned encoders. During each MPPI rollout, observed obstacle points are transformed into the predicted body frame and evaluated against the footprint. All operations are batched in JAX, leveraging GPU parallelism for real-time receding-horizon control. Experiments show that \algname{} accelerates batched distance evaluation over a learned point-to-robot baseline, preserves feasible motion where convex-footprint planners fail, and remains robust under dense static and moving obstacles. The same framework deploys on differential-drive, Ackermann, omnidirectional, and hybrid-mode platforms by changing only the footprint description and motion model without per-platform training. Pairing exact footprint geometry with sampling-based predictive control thus offers a practical, training-free path to footprint-aware local navigation across diverse robots.

\end{abstract}


\section{Introduction}

\begin{figure*}[!t]
    \centering
    \includegraphics[width=0.9\textwidth]{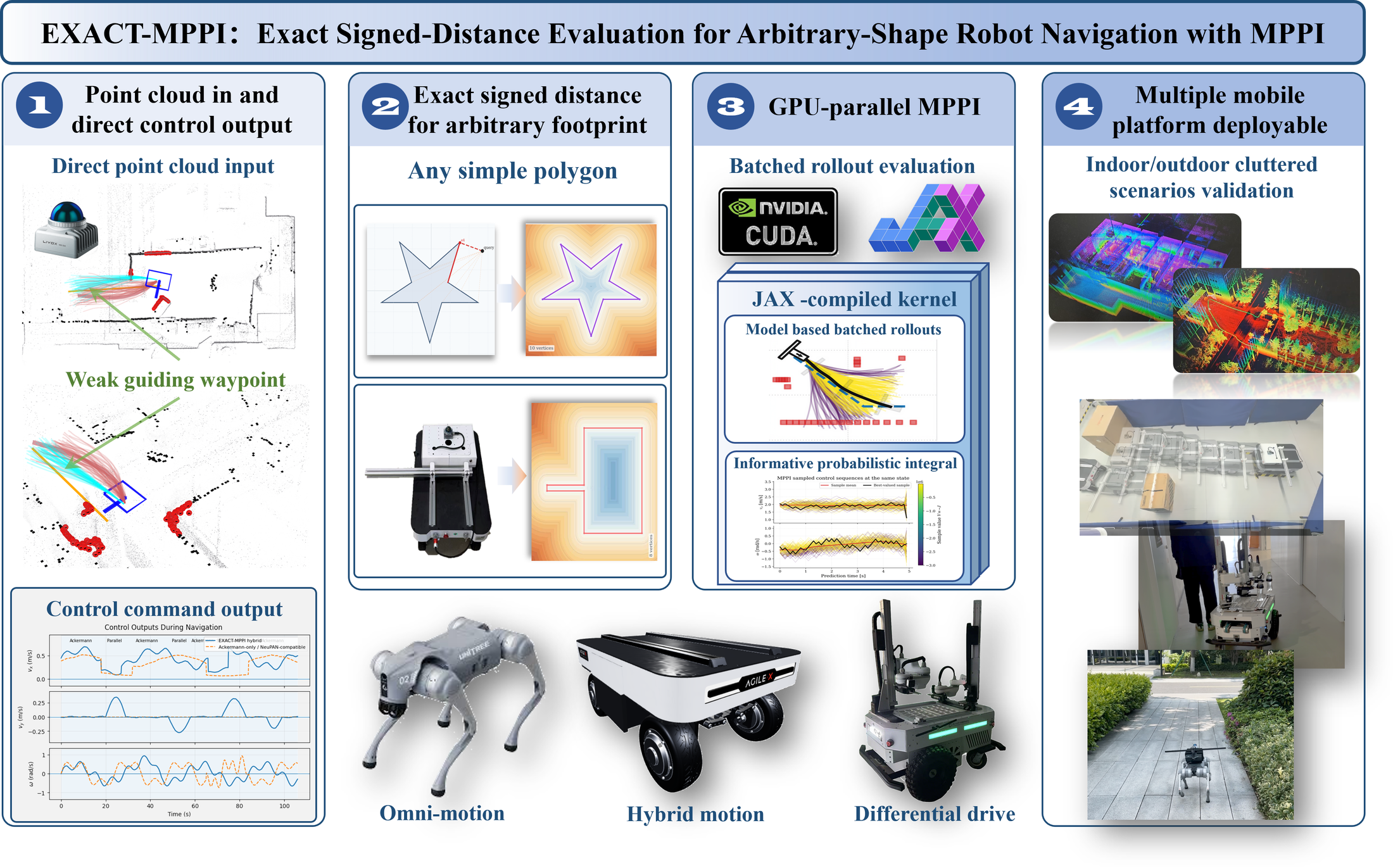}
    \caption{Overview of \algname{}, illustrating the four main contributions: (1) perception-to-control navigation from direct point clouds and weak waypoint guidance to motion commands; (2) analytic, exact signed-distance evaluation for 2D footprints, including non-convex polygons; (3) a GPU-parallel MPPI controller with JAX-compiled batched rollouts; and (4) cross-platform validation on differential-drive, omnidirectional, and hybrid robots in indoor and outdoor cluttered environments.}
    \label{fig:framework_overview}
\end{figure*}

Ground robots are increasingly deployed in cluttered environments, where safe and efficient local navigation requires careful reasoning about the robot's geometry. Many such robots have intricate planar footprints; for example, forklifts carrying palletized loads in warehouse aisles \cite{vivaldini2010robotic}, agricultural platforms towing implements through orchard rows \cite{peng2024optimization}, and mobile manipulators transporting objects through constrained workspaces \cite{kennel2024payload}. The resulting footprint is often non-convex and irregular. In cluttered environments, the available clearance is frequently comparable to the geometric detail of the robot footprint, and accurately accounting for this geometry becomes essential for finding feasible motions.

Navigating such robots in cluttered scenes presents several technical challenges. First, accurately evaluating collision geometry against a complex footprint is computationally expensive. To meet real-time requirements, many planners approximate the robot footprint by circles, rectangles, or convex hulls. These approximations can be effective in open environments, but in narrow passages they inflate the collision boundary and eliminate maneuvers that the true footprint would have allowed. Second, most navigation pipelines construct an intermediate environment representation (e.g., an occupancy grid or costmap) before planning can proceed, adding latency and introducing sensitivity to design choices such as grid resolution and inflation radius. Third, navigation commands must respect the kinematic constraints of the robot (e.g., differential-drive or Ackermann), and the planning framework must accommodate different motion models without redesign of its core machinery. Fourth, learning-based approaches that aim to circumvent these issues are typically tied to a specific robot platform and environment, and transferring them to new settings requires extensive retraining in simulation.

To address these challenges, we propose \algname{} -- a training-free, perception-to-control local navigation framework for ground robots with complex 2D footprints operating in cluttered environments. The framework computes the exact minimum signed distance analytically between observed obstacle points and the robot footprint, which is modeled as a simple polygon that can be convex or concave, without convex decomposition or inflation. Motion commands are generated directly from point clouds perceived by onboard LiDAR, without constructing any intermediate environment representation such as an occupancy grid, signed-distance field, or local costmap. Collision-aware and smooth local navigation is achieved through Model Predictive Path Integral (MPPI) control~\cite{williams2017model}, a sampling-based predictive controller that does not require gradient information or differentiable cost functions and supports a wide range of robot kinematics. MPPI is driven by a weak guidance path, and the signed-distance evaluation is incorporated into the cost function as a safety penalty. The signed-distance evaluator and the MPPI trajectory rollouts are implemented as batched JAX computations~\cite{jax2018github} with GPU acceleration, enabling real-time planning over a receding horizon while preserving the exact polygonal footprint.

The main contributions of this work, illustrated in Fig.~\ref{fig:framework_overview}, are as follows:
\begin{itemize}
    \item We propose \algname{}, a perception-to-control local navigation framework that maps local LiDAR point-cloud observations and weak waypoint guidance directly to motion commands, without constructing occupancy grids, signed-distance fields, or local costmaps.

    \item We develop an analytic point-to-footprint signed-distance evaluator for explicit 2D footprint representations. The simple-polygon formulation handles convex and concave planar footprints in a unified manner, while a rectangle-cover specialization accelerates evaluation for rectilinear footprints.

    \item We integrate this signed-distance evaluator into MPPI as a footprint-aware safety cost and validation mechanism. The evaluator and rollout computations are implemented as JAX-compiled batched operations, parallelized across rollout samples, horizon steps, and obstacle points.

    \item We demonstrate that the same collision-evaluation and MPPI control structure transfers across multiple ground-robot platforms with different footprints and motion models (differential-drive, Ackermann-steering, omnidirectional, and hybrid). Deploying \algname{} on a new platform requires only specifying its footprint polygon and motion model. Experiments span indoor environments (office, hallway) and outdoor environments (garden-like) with both static and low-speed dynnamic obstacles, in simulation and on real hardware.
\end{itemize}

The rest of the paper is organized as follows. Section~\ref{sec:related_work} reviews related work, and Section~\ref{sec:preliminary} introduces preliminary concepts and notation. Section~\ref{sec:methodology} describes the exact minimum signed-distance evaluation and the MPPI cost formulation. Section~\ref{sec:experiments_and_results} presents the experimental design and results. Section~\ref{sec:conclusion} concludes the paper.

\section{Related Work}
\label{sec:related_work}
This section reviews three areas of related work: modular map-based navigation (Sec.~\ref{subsec:related_classical}) and learning-based navigation (Sec.~\ref{subsec:related_learning}), which together cover the dominant approaches to local navigation. Sec.~\ref{subsec:related_mppi} then reviews sampling-based predictive control, the control framework adopted in this work.

\subsection{Modular Map-Based Navigation}
\label{subsec:related_classical}
Most robot navigation systems follow a modular pipeline of perception, mapping, planning, and control. Sensor data is first converted into a structured environment representation such as an occupancy grid, costmap, or Euclidean Signed Distance Field (ESDF)~\cite{oleynikova2017voxblox}, in which obstacle observations are rasterized into occupied cells or pre-computed distance values, and the robot footprint is approximated for collision checking. A local planner then generates kinematically feasible, collision-free trajectories on this representation in response to nearby and possibly dynamic obstacles, and a low-level controller tracks the resulting motion.

Local planners operating on map representations generally fall into four families. \emph{Reactive methods}, including artificial potential fields~\cite{khatib1986real}, vector field histograms~\cite{borenstein1991vector}, the nearness diagram method~\cite{minguez2004nearness}, and the dynamic window approach~\cite{fox1997dynamic}, compute steering commands from local sensor data while reducing the robot to a point or a disc. \emph{Sampling-based methods}, including lattice planners~\cite{mcnaughton2011motion} and Falco~\cite{zhang2020falco}, sample candidate motions and score them against a cost function. \emph{Search-based methods}, including Hybrid $A^*$~\cite{dolgov2010path}, discretize the configuration space and apply heuristic search over a costmap. \emph{Optimization-based methods}, including the Timed Elastic Band~\cite{rosmann2017kinodynamic}, EGO-Planner~\cite{zhou2020ego}, corridor-based planners that decompose free space into convex regions~\cite{ding2021epsilon,han2023efficient}, and recent swept-volume extensions~\cite{li2025efficient,wang2024implicit}, generate trajectories by optimizing a cost function over a parameterized trajectory representation.

A closely related body of work embeds collision avoidance within a model predictive control (MPC) framework, solving a finite-horizon optimal control problem at each control step. Representative examples include model predictive contouring control~\cite{brito2019model}, CIAO~\cite{schoels2020nmpc}, and optimization-based collision avoidance (OBCA)~\cite{zhang2020optimization}, which handles full-shape robot-obstacle distance through a convex-duality reformulation. Control barrier functions~\cite{ames2019control,long2021learning} provide a complementary, control-level safety mechanism by enforcing forward invariance of a safe set defined by a smooth distance-like function.

These methods share three limitations when applied to robots with complex footprints in cluttered environments. First, they rely on an intermediate map representation whose discrete structure introduces sensitivity to grid resolution and inflation radius~\cite{oleynikova2017voxblox,zhou2020ego} and a loss of geometric detail relative to the raw sensor data. Second, obstacle and robot geometry are typically simplified into convex, smooth, or inflated proxies~\cite{fox1997dynamic,rosmann2017kinodynamic,brito2019model}, inflating the collision boundary and eliminating maneuvers that the true footprint would have permitted. Methods that retain exact geometry, such as OBCA~\cite{zhang2020optimization} and swept-volume extensions~\cite{li2025efficient,wang2024implicit}, can incur a computational cost that limits real-time use on densely cluttered problems. Third, when the footprint is geometrically detailed, the modular separation between perception, planning, and control compounds the first two limitations: small geometric details available in the raw sensor data are abstracted away at the mapping stage and are no longer accessible to the planner~\cite{han2025neupan}. \algname{} addresses these limitations by evaluating analytic point-to-footprint signed distances directly between observed obstacle points and an explicit planar footprint representation in the robot body frame. It avoids map rasterization in the local collision-evaluation loop and generates control commands through a perception-to-control sampling-based predictive controller.

\subsection{Learning-Based Navigation}
\label{subsec:related_learning}
A second body of work replaces parts or all of the navigation pipeline with learned components. We distinguish two groups: end-to-end policies that map sensor observations directly to control, and model-based learning approaches that retain a model-based planner or controller but learn one or more of its subroutines.

\textit{End-to-end policies.}
Pioneered by ALVINN~\cite{pomerleau1988alvinn} and revived for highway driving by Bojarski et al.~\cite{bojarski2016end}, end-to-end methods have demonstrated strong performance in automotive domains through large-scale imitation learning~\cite{chen2024end}. Recent work has also explored LiDAR-based reinforcement learning policies for navigation in dynamic environments~\cite{zhu2024learn}. To improve sample efficiency and generalization, subsequent work has introduced inductive biases for planning, including value iteration networks~\cite{tamar2016value} and motion-planning networks such as MPNet~\cite{qureshi2019motion}. Recent foundation-model-style navigation policies such as ViNT~\cite{shah2023vint} further target cross-embodiment generalization across heterogeneous robots and environments. Despite this progress, end-to-end methods face a structural data requirement. Unlike autonomous driving, where fleet-scale data is abundant, applications such as warehousing and orchard operations are characterized by data scarcity, distributional shifts, and platform-specific footprint geometries. Policies that perform well on benchmark distributions frequently exhibit out-of-distribution degradation when environment topology, obstacle density, or robot dimensions differ from the training set, often necessitating site-specific retraining or parameter tuning.

\textit{Model-based learning approaches.}
A second group retains a model-based formulation but replaces specific subroutines with learned components, aiming to preserve interpretability and constraint handling while accelerating expensive computations. Hybrid architectures such as iPlanner~\cite{yang2023iplanner} combine a learned perception front-end with classical trajectory optimization. The most closely related example, and the most direct point of comparison for our work, is NeuPAN~\cite{han2025neupan}. NeuPAN formulates navigation as an end-to-end model-based learning problem in which point-cloud observations, robot geometry, and predictive control are coupled within a unified perception-to-control pipeline. A deep unfoldered neural encoder (DUNE) maps raw obstacle points to latent distance features, which are then consumed by a neural regularized motion planner (NRMP). The encoder amortizes the cost of solving a per-point convex distance subproblem derived from a duality-based formulation of point-to-body distance~\cite{zhang2020optimization}, enabling real-time evaluation across thousands of points. NeuPAN demonstrates strong performance in cluttered environments on differential-drive, wheel-legged, and passenger-vehicle platforms.

\algname{} shares NeuPAN's robot-centric, perception-to-control viewpoint and its goal of avoiding intermediate map representations, but differs in three design choices. First, NeuPAN's DUNE encodes the robot footprint into network weights through training, so the encoder must be retrained whenever the footprint changes. In contrast, we compute the point-to-footprint signed distance analytically, so adapting to a new footprint requires only updating the polygon description. Second, NeuPAN's distance formulation models the robot as a convex set, with non-convex bodies represented as the union of convex sets~\cite{wei2026neural}; our analytic evaluator handles simple polygons, including concave ones, as a single object. Third, NeuPAN's NRMP solves a biconvex motion-planning problem via proximal alternating minimization, a gradient-based scheme whose regularizer parameters are tuned in simulation. Because the analytic distance function we use is non-smooth, we instead adopt sampling-based MPPI control, which is gradient-free, batches naturally on GPU, and accommodates a wide range of robot kinematics without retraining.

\subsection{Sampling-Based Predictive Control}
\label{subsec:related_mppi}
\algname{} adopts Model Predictive Path Integral (MPPI) control~\cite{williams2017model,williams2018information}, the most widely used formulation of sampling-based predictive control. In this family, trajectory candidates are drawn from a stochastic distribution and weighted by their costs. This formulation accommodates non-smooth, non-convex cost functions and a broad class of rollout dynamics, and its sampling structure maps naturally onto GPU parallelism. MPPI has been deployed for aggressive off-road driving~\cite{williams2016aggressive}, autonomous racing, and agile flight~\cite{minavrik2024model}. Subsequent work has targeted robustness and sample efficiency: Log-MPPI~\cite{mohamed2022autonomous} uses a heavy-tailed sampling distribution for broader trajectory-space exploration, and smooth variants~\cite{kim2022smooth} reduce control chatter.

Despite these advances, MPPI's treatment of collision avoidance has received comparatively less attention. Many implementations delegate collision evaluation to map-based primitives such as costmap lookups~\cite{macenski2020marathon2} or grid-based signed-distance fields~\cite{oleynikova2017voxblox}. These primitives inherit the footprint-simplification and resolution-sensitivity issues discussed in Sec.~\ref{subsec:related_classical}, particularly when the footprint is non-convex or obstacle density is high.

To address this gap, \algname{} integrates MPPI with an analytic body-frame signed-distance evaluator that operates directly on observed obstacle points and the robot's polygonal footprint. The evaluator and rollout computations are implemented as JAX-compiled batched operations. This combines MPPI's sampling parallelism with exact-shape collision evaluation, without any costmap or ESDF queries.

\section{Preliminaries}
\label{sec:preliminary}

This section defines the inputs, geometry, signed-distance quantity, kinematic models, and MPPI formulation that the proposed \algname{} builds on.

\subsection{Problem Setting}
\label{subsec:scope_and_assumptions}
At each replanning instant, the controller receives a local point cloud from onboard LiDAR and the current robot state from onboard state estimation. We denote the current planar pose of the robot by $\vect{q}_0 = [x_0, y_0, \theta_0]^\top$, expressed in the local planning frame. Collision costs are evaluated with respect to the observed local obstacle points after preprocessing, rather than against a continuous obstacle surface or a rasterized map. The collision-relevant geometry of the robot is captured by a two-dimensional projected footprint that includes the chassis together with any rigidly attached or carried geometry. The detailed model is given in Sec.~\ref{subsec:effective_footprint}.

We assume that weak guidance is available from an upstream module. Examples include a target pose, waypoint sequence, reference path, or learned navigation prior~\cite{shah2023gnm}. This assumption matches the applications that motivate this work: warehouse forklifts receive target pallet poses from fleet- or task-level planners~\cite{vivaldini2010robotic}, agricultural platforms follow between-row reference paths generated from prior maps or row-detection modules~\cite{peng2024optimization}, and mobile manipulators receive target object poses from task planners~\cite{kennel2024payload}. In all these cases, the local controller is responsible for following the high-level intent safely through clutter, not for generating the intent itself.

To handle dynamic obstacles, the controller treats them as quasi-static within a single short MPPI horizon. Collisions are then re-evaluated at each replanning step through receding-horizon updates. The core formulation does not include an explicit motion-prediction model, but one can be added separately if required.

\subsection{Robot Effective Footprint}
\label{subsec:effective_footprint}
We represent the collision-relevant geometry of the robot by an \emph{effective footprint}: a compact planar set in the robot body frame,
\begin{equation}
    \mathcal{B}_{\mathrm{eff}} \subset \mathbb{R}^2,
\end{equation}
used for collision checking and clearance evaluation during local navigation. Depending on the platform, $\mathcal{B}_{\mathrm{eff}}$ may correspond to the chassis alone or include any rigidly attached or carried geometry. We write
\begin{equation}
    \mathcal{B}_{\mathrm{eff}}
    =
    \mathcal{B}_{\mathrm{chassis}}
    \cup
    \mathcal{B}_{\mathrm{add}},
\end{equation}
where $\mathcal{B}_{\mathrm{chassis}}$ denotes the nominal chassis and $\mathcal{B}_{\mathrm{add}}$ denotes any added geometry, such as a palletized load on a forklift, a trailing implement on an agricultural platform, or an object grasped by a mobile manipulator.

The effective footprint is assumed fixed in the body frame during a navigation episode. Different platform configurations are accommodated by updating $\mathcal{B}_{\mathrm{eff}}$, with no per-platform training required. In practice, $\mathcal{B}_{\mathrm{eff}}$ is represented either by a rectangle cover or by a (possibly non-convex) simple polygon, depending on the platform geometry. The computational details are given in Sec.~\ref{subsec:footprint_modeling}.

\subsection{Robot--Obstacle Signed Distance}
\label{subsec:signed_distance_quantity}
At each replanning instant, the controller uses a preprocessed local obstacle set
\begin{equation}
    \mathcal{O} = \{\vect{o}_{i}\}_{i=1}^{N},
    \qquad
    \vect{o}_{i} \in \mathbb{R}^2,
\end{equation}
expressed in the current planning frame. Here $N$ denotes the number of obstacle points supplied to the controller after height filtering and downsampling, rather than the number of original LiDAR returns.

The signed distance from a body-frame query point $\vect{p}$ to the effective footprint, denoted
\begin{equation}
    d^{\pm}(\vect{p}, \mathcal{B}_{\mathrm{eff}}),
    \label{eq:signed_distance}
\end{equation}
is negative inside $\mathcal{B}_{\mathrm{eff}}$, zero on its boundary, and positive outside. Throughout this paper, the query is taken against observed obstacle points rather than a continuous obstacle surface. Explicit formulas for $d^{\pm}$ are derived in Sec.~\ref{subsec:distance_evaluation}.

The minimum signed distance from $\mathcal{O}$ to the robot footprint at the current pose is
\begin{equation}
    d^{\min}
    =
    \min_{i=1,\dots,N}
    d^{\pm}\!\left(
    \vect{o}_{i},
    \mathcal{B}_{\mathrm{eff}}
    \right).
\end{equation}
A negative value $d^{\min}<0$ indicates that at least one obstacle point lies inside the effective footprint. A positive value is the minimum point-wise clearance to the local obstacle set.

\subsection{Robot Kinematic Models}
\label{subsec:motion_models}

Let $m \in \mathcal{M}$ index the motion model used for state propagation. The planar pose $\vect{q}_h = [x_h, y_h, \theta_h]^\top$ is propagated with a forward-Euler discretization,
\begin{equation}
    \vect{q}_{h+1}
    =
    \vect{q}_{h}
    +
    \mathcal{F}_m(\vect{q}_{h},\vect{u}_{h})\,\Delta t,
\end{equation}
where $\vect{u}_{h} \in \mathbb{R}^{n_u}$ is the control input, $\Delta t$ is the sampling interval, and $\mathcal{F}_m(\cdot)$ denotes the model-specific continuous-time kinematics.

We consider the following motion models in our experiments. The differential-drive (or unicycle) model takes control input $\vect{u}=[v,\omega]^\top$, where $v$ is the linear velocity and $\omega$ is the angular velocity, with kinematics
\begin{equation}
    \mathcal{F}_{\mathrm{diff}}(\vect{q}, \vect{u})
    =
    \begin{bmatrix}
        v \cos \theta \\
        v \sin \theta \\
        \omega
    \end{bmatrix}.
\end{equation}
The Ackermann (or bicycle) model takes $\vect{u}=[v,\delta]^\top$, where $\delta$ is the steering angle, with kinematics
\begin{equation}
    \mathcal{F}_{\mathrm{ack}}(\vect{q}, \vect{u})
    =
    \begin{bmatrix}
        v \cos \theta \\
        v \sin \theta \\
        v / L \tan \delta
    \end{bmatrix},
\end{equation}
where $L$ is the wheelbase.

For platforms that accept planar body-frame velocity commands, the omni-motion model is
\begin{equation}
    \mathcal{F}_{\mathrm{omni}}(\vect{q}, \vect{u})
    =
    \begin{bmatrix}
        v_x \cos \theta - v_y \sin \theta \\
        v_x \sin \theta + v_y \cos \theta \\
        \omega
    \end{bmatrix},
\end{equation}
where $v_x$ and $v_y$ are the body-frame longitudinal and lateral velocities. The spin-in-place mode is represented by
\begin{equation}
    \mathcal{F}_{\mathrm{spin}}(\vect{q}, \vect{u})
    =
    \begin{bmatrix}
        0 \\
        0 \\
        \omega
    \end{bmatrix}.
\end{equation}
For hybrid platforms that additionally support lateral translation (e.g., the AgileX Ranger Mini~\cite{agilex_ranger_mini}), a parallel-motion mode is modeled as
\begin{equation}
    \mathcal{F}_{\mathrm{para}}(\vect{q}, \vect{u})
    =
    \begin{bmatrix}
        -v_{\mathrm{para}} \sin \theta \\
        v_{\mathrm{para}} \cos \theta  \\
        0
    \end{bmatrix},
\end{equation}
where $v_{\mathrm{para}}$ is the signed lateral body-frame velocity, that is, translation along the body-frame $y$-axis with the heading preserved.

\subsection{MPPI Formulation}
\label{subsec:mppi_problem}

Model Predictive Path Integral (MPPI) control~\cite{williams2017model} is used as the receding-horizon local controller. At each replanning instant, the controller maintains a nominal control sequence
\begin{equation}
    \mathbb{U}
    =
    \{\vect{u}_{0},\dots,\vect{u}_{T-1}\},
\end{equation}
where $T$ is the planning horizon length. MPPI generates $K$ sampled rollout sequences by perturbing this nominal sequence,
\begin{equation}
    \vect{u}_{h}^{(r)}
    =
    \vect{u}_{h}
    +
    \boldsymbol{\epsilon}_{h}^{(r)},
    \qquad
    h=0,\dots,T-1,\quad r=1,\dots,K,
\end{equation}
where $\boldsymbol{\epsilon}_{h}^{(r)}$ is the sampled control perturbation for rollout $r$ at horizon step $h$. The corresponding perturbed control sequence is denoted by
\begin{equation}
    \mathbb{U}^{(r)}
    =
    \{\vect{u}_{0}^{(r)},\dots,\vect{u}_{T-1}^{(r)}\}.
\end{equation}
Using the Euler discretization in Sec.~\ref{subsec:motion_models}, each rollout starts from $\vect{q}_0$ and follows
\begin{equation}
    \vect{q}_{h+1}^{(r)}
    =
    \vect{q}_{h}^{(r)}
    +
    \mathcal{F}_m
    \left(
    \vect{q}_{h}^{(r)},
    \vect{u}_{h}^{(r)}
    \right)\Delta t,
    \qquad
    \vect{q}_{0}^{(r)}=\vect{q}_0 ,
\end{equation}
under the selected motion model $m$.

At a high level, MPPI evaluates sampled control sequences using a finite-horizon cost that combines task, control-regularization, and safety terms. In \algname{}, the safety term is defined by the minimum signed distance between the predicted robot footprint and the local obstacle observation. For rollout $r$, we write
\begin{equation}
    \begin{aligned}
        J^{(r)}
        =
        \sum_{h=0}^{T-1}
        \Big[
         & \phi_{\mathrm{task}}(\vect{q}_{h}^{(r)},\vect{u}_{h}^{(r)})
            +
        \phi_{\mathrm{ctrl}}(\vect{u}_{h}^{(r)})                       \\
         & +
            \phi_{\mathrm{obs}}(d_{h}^{\min,(r)})
            \Big],
    \end{aligned}
\end{equation}
where $\phi_{\mathrm{task}}$ represents navigation objectives such as goal seeking, path following, or progress along a reference path, $\phi_{\mathrm{ctrl}}$ regularizes the sampled control command, and $\phi_{\mathrm{obs}}$ penalizes collision or insufficient clearance. The scalar $d_{h}^{\min,(r)}$ denotes the minimum signed distance evaluated at the rollout state $\vect{q}_{h}^{(r)}$ over the local obstacle points; its rollout-frame definition is given in Sec.~\ref{subsec:distance_evaluation}.

After all rollouts are evaluated, MPPI assigns an importance weight to each rollout according to its relative cost. Let
\begin{equation}
    \beta = \min_{r=1,\dots,K} J^{(r)} .
\end{equation}
The normalized weight of rollout $r$ is
\begin{equation}
    \omega^{(r)}
    =
    \frac{
        \exp\left(-(J^{(r)}-\beta)/\lambda\right)
    }{
        \sum_{j=1}^{K}
        \exp\left(-(J^{(j)}-\beta)/\lambda\right)
    },
\end{equation}
where $\lambda>0$ is the temperature parameter. The nominal control sequence is then updated by the weighted average of the sampled perturbations,
\begin{equation}
    \vect{u}_{h}
    \leftarrow
    \vect{u}_{h}
    +
    \sum_{r=1}^{K}
    \omega^{(r)}
    \boldsymbol{\epsilon}_{h}^{(r)},
    \qquad
    h=0,\dots,T-1 .
\end{equation}
This update is a stochastic receding-horizon improvement step rather than an exact solution of the nonlinear finite-horizon optimal control problem. After the update, only the first control command is executed. The horizon is then shifted forward, the local obstacle observation is refreshed, and the procedure is repeated at the next replanning instant. The specific signed-distance computation, safety cost, feasibility screening, trajectory validation, and batched implementation used in \algname{} are described in Sec.~\ref{sec:methodology}.

The main MPPI parameters that determine the sampling budget are the rollout count $K$, horizon length $T$, integration step $\Delta t$, control-perturbation distribution, and temperature $\lambda$. The physical prediction horizon is $T\Delta t$. Each control cycle samples $K$ candidate control sequences, corresponding to $KT$ perturbed control inputs and $K(T+1)$ rollout states before the first command is executed. In the footprint-aware setting, the obstacle point budget $N$, safety margin $d_{\mathrm{safe}}$, and cost weights further determine the number and influence of collision checks performed per cycle.

\section{Methodology}
\label{sec:methodology}
\begin{figure}[t]
    \centering
    \includegraphics[width=\columnwidth]{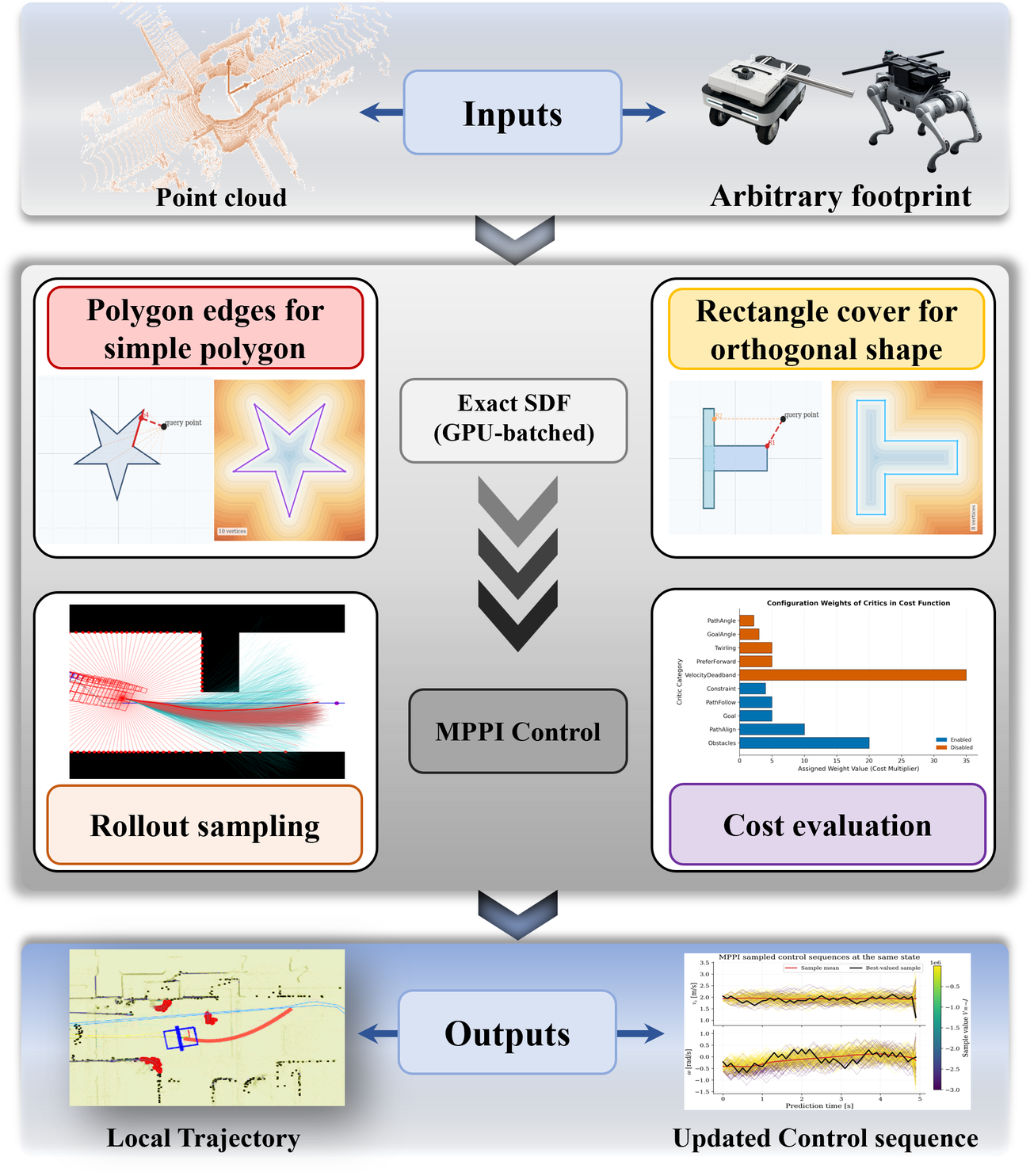}
    \caption{System architecture of \algname{}.}
    \label{fig:pipeline_architecture}
\end{figure}
We describe the computational realization of \algname{}, building on the formulation introduced in Sec.~\ref{sec:preliminary}. As shown in Fig.~\ref{fig:pipeline_architecture}, local point-cloud observations are reduced to a fixed-budget obstacle set. MPPI rollouts are propagated under the selected kinematics, obstacle points are transformed into each predicted body frame, and signed-distance values are used for rollout scoring, control-sequence update, and safety validation.

\subsection{Overview of \algname{}}
\label{subsec:method_overview}

At each replanning instant, \algname{} receives three inputs: weak guidance from an upstream module, the current robot velocity, and a compact local obstacle set. The weak guidance is provided as a target pose or short waypoint sequence in the current robot frame, which defines the local navigation intent without requiring a dense reference trajectory. The current chassis velocity, obtained from odometry, is used to warm-start the nominal control sequence. The obstacle set is obtained from the local point cloud by retaining points within a task-relevant height range and downsampling them to a fixed budget \(N\). The downsampling procedure is applied before rollout evaluation and determines the point-wise obstacle set against which clearance is measured.

Given these inputs, MPPI samples $K$ control rollouts over $T$ horizon steps, and propagates the corresponding state trajectories under the selected motion model. For each predicted state, the $N$ obstacle points are transformed into the predicted robot body frame, where signed distance is evaluated against the explicit 2D effective footprint. The resulting minimum signed-distance values are used with task and control costs to score the rollouts, screen unsafe candidates, update the nominal control sequence, and validate the selected trajectory before executing only the first command.

\subsection{Computational Footprint Representations}
\label{subsec:footprint_modeling}

Starting from the effective-footprint model in Sec.~\ref{subsec:effective_footprint}, we use two representations for signed-distance evaluation. The choice between them reflects a deliberate trade-off between geometric generality and computational efficiency. The general simple-polygon representation handles arbitrary boundary shapes, including non-convex footprints, but requires per-edge projection and a point-in-polygon test. The rectangle-cover representation is restricted to orthogonal footprints, but admits a closed-form point-to-box distance that is substantially cheaper to batch on the GPU. \algname{} retains both routes so that rectilinear platforms benefit from the faster evaluator without sacrificing the ability to handle more general footprints. We use $R$ for the number of rectangles in a rectangle cover and $B$ for the number of boundary edges (equivalently, vertices) in a polygonal footprint.

\subsubsection{Orthogonal Footprints via Rectangle Covering}
A footprint is termed \emph{rectilinear} when its boundary edges are axis-aligned and meet at right angles. Many vehicle chassis and attached implements fall in this class. For such footprints, the effective footprint is represented as a union of axis-aligned rectangles in the body frame,
\begin{equation}
    \mathcal{B}_{\mathrm{eff}}
    =
    \bigcup_{j=1}^{R} \mathcal{R}_j .
\end{equation}
Each rectangle $\mathcal{R}_j$ is parameterized by its center $\vect{c}_j \in \mathbb{R}^2$ and half-extent $\vect{s}_j \in \mathbb{R}_{+}^2$. We retain this representation alongside the general polygon route purely for computational efficiency. The closed-form point-to-box distance derived in Sec.~\ref{subsec:distance_evaluation} requires only elementwise arithmetic and a single vector norm per rectangle, with no branching, sorting, or topology test, which makes it well suited to batched rollout evaluation on the GPU.

\subsubsection{General Simple Polygonal Footprints}

For footprints whose boundary is not rectilinear, including concave shapes such as those resulting from attached implements or grasped objects, the effective footprint is described by an ordered cyclic sequence of $B$ vertices in the body frame,
\begin{equation}
    \mathcal{V}
    =
    (\vect{v}_1,\,\vect{v}_2,\,\dots,\,\vect{v}_B),
    \qquad
    \vect{v}_b \in \mathbb{R}^2,
\end{equation}
with the wrap-around convention $\vect{v}_{B+1} \equiv \vect{v}_1$. The polygon is \emph{simple} in the sense that its boundary is a non-self-intersecting closed curve. The boundary consists of $B$ directed edges, each represented as a vector,
\begin{equation}
    \vect{e}_b
    =
    \vect{v}_{b+1} - \vect{v}_b,
    \qquad
    b=1,\dots,B.
\end{equation}
This vector form is used as the basic primitive in the point-to-segment distance derived in Sec.~\ref{subsec:distance_evaluation}.

\subsection{Signed Distance Evaluation}
\label{subsec:distance_evaluation}
We now instantiate the signed-distance evaluation in Eq.~\eqref{eq:signed_distance}. For both footprint representations in Sec.~\ref{subsec:footprint_modeling}, distance queries are evaluated in the predicted robot body frame. Thus, the effective footprint remains fixed during rollout evaluation, while the preprocessed obstacle points are transformed according to each predicted rollout pose.

\subsubsection{Rollout-Frame Transformation}
Let $\mathcal{O}=\{\vect{o}_i\}_{i=1}^{N}$ denote the preprocessed local obstacle set supplied to the controller, expressed in the current planning frame. During rollout evaluation, this same obstacle set is copied across sampled trajectories and horizon steps. For rollout $r$ and horizon step $h$, let
\begin{equation}
    \vect{q}_{h}^{(r)}
    =
    [x_{h}^{(r)},\ y_{h}^{(r)},\ \theta_{h}^{(r)}]^\top
\end{equation}
denote the predicted robot pose relative to the current planning frame, with
$\vect{t}_{h}^{(r)}=[x_{h}^{(r)},\ y_{h}^{(r)}]^\top$.
Each obstacle point is re-expressed in the predicted body frame as
\begin{equation}
    \vect{p}_{h,i}^{b,(r)}
    =
    \mathbf{R}\!\left(\theta_{h}^{(r)}\right)^\top
    \left(
    \vect{o}_i
    -
    \vect{t}_{h}^{(r)}
    \right),
    \quad
    i=1,\dots,N .
\end{equation}
The resulting points are then evaluated against the effective footprint in the predicted body frame. In this way, \algname{} keeps the footprint representation static and instead transforms the local obstacle observation into the body frame of each rollout pose for signed-distance evaluation.

\subsubsection{Distance to Orthogonal Footprints}
For a body-frame query point $\vect{p}$ and a rectangle $\mathcal{R}_j$ with center $\vect{c}_j$ and half-extent $\vect{s}_j$, define
\begin{equation}
    \vect{a}_j(\vect{p})
    =
    |\vect{p}-\vect{c}_j|-\vect{s}_j ,
\end{equation}
where the absolute value is applied elementwise. The point-to-box signed distance is
\begin{equation}
    d^{\pm}_{\mathrm{box},j}(\vect{p})
    =
    \|\max(\vect{a}_j(\vect{p}),0)\|_2
    +
    \min\left(
    \max(a_{j,x},a_{j,y}),0
    \right),
\end{equation}
where $\max(\cdot,0)$ inside the norm is taken elementwise and $a_{j,x}, a_{j,y}$ are the components of $\vect{a}_j(\vect{p})$. For a rectangle-cover footprint, the signed distance is obtained by reduction over the $R$ rectangles,
\begin{equation}
    d^{\pm}_{\mathrm{rect}}
    \left(
    \vect{p},
    \mathcal{B}_{\mathrm{eff}}
    \right)
    =
    \min_{j=1,\dots,R}
    d^{\pm}_{\mathrm{box},j}(\vect{p}) .
\end{equation}

\subsubsection{Distance to General Simple Polygonal Footprint}
For a simple polygonal footprint, the unsigned distance to the boundary is the minimum distance to its polygon edges. For an edge
$\vect{e}_b=\vect{v}_{b+1} - \vect{v}_b$,
define the clipped projection parameter
\begin{equation}
    \alpha_b(\vect{p})
    =
    \max
    \left(
    0,
    \min
    \left(
        1,
        \frac{
                (\vect{p}-\vect{v}_b)^\top
                (\vect{v}_{b+1}-\vect{v}_b)
            }{
                \|\vect{e}_b\|^2
            }
        \right)
    \right),
\end{equation}
and the corresponding point-to-segment distance
\begin{equation}
    d_{\mathrm{seg},b}(\vect{p})
    =
    \left\|
    \vect{p}
    -
    \left(
    \vect{v}_b
    +
    \alpha_b(\vect{p})
    \right)
    \vect{e}_b
    \right\|_2 .
\end{equation}
The sign is assigned using a point-in-polygon test (implemented via ray casting),
\begin{equation}
    \sigma(\vect{p},\mathcal{B}_{\mathrm{eff}})
    =
    \begin{cases}
        -1, & \vect{p} \in \mathcal{B}_{\mathrm{eff}},    \\
        +1, & \vect{p} \notin \mathcal{B}_{\mathrm{eff}}.
    \end{cases}
\end{equation}
The signed distance to the simple polygonal footprint is then
\begin{equation}
    d^{\pm}_{\mathrm{poly}}
    \left(
    \vect{p},
    \mathcal{B}_{\mathrm{eff}}
    \right)
    =
    \sigma
    \left(
    \vect{p},
    \mathcal{B}_{\mathrm{eff}}
    \right)
    \min_{b=1,\dots,B}
    d_{\mathrm{seg},b}(\vect{p}) .
\end{equation}
This quantity is the exact signed distance for the represented simple polygon, whether convex or non-convex.

\begin{figure}[t]
    \centering
    \includegraphics[width=0.95\columnwidth]{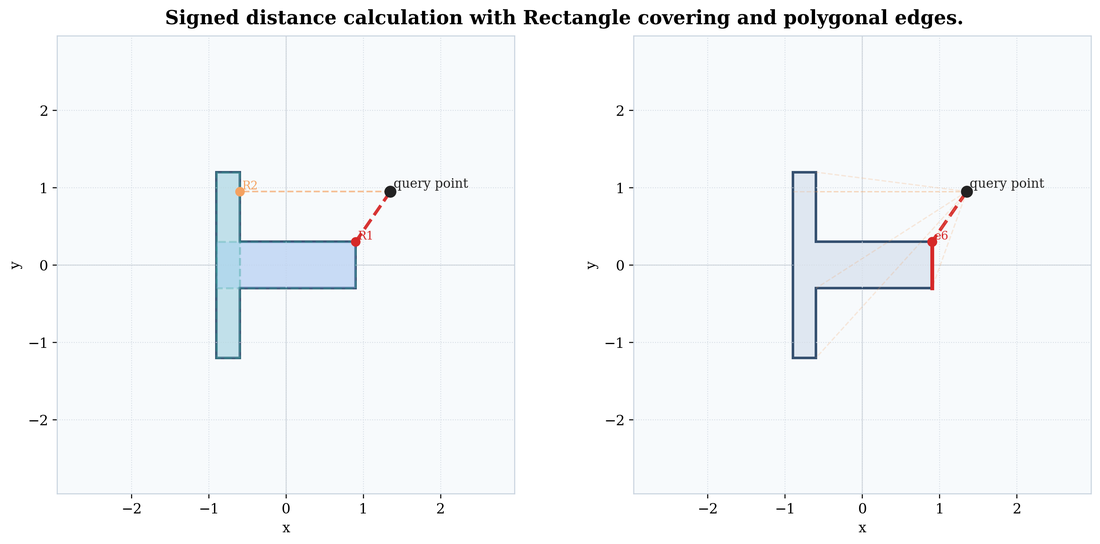}
    \caption{Distance evaluation for an orthogonal footprint: rectangle cover (left) versus polygon edges (right).}
    \label{fig:t_shape_distance_methods}
\end{figure}

Figure~\ref{fig:t_shape_distance_methods} compares the two evaluators on a T-shaped rectilinear footprint with an exterior query point. The rectangle-cover route (left) takes the minimum over per-rectangle point-to-box distances, while the polygon-edge route (right) takes the minimum over per-edge point-to-segment distances. Both identify the same nearest boundary point and return the same exterior signed distance, which is the regime that MPPI actually uses for safety penalization.

\textbf{Exactness and sign correctness.}
For a simple polygonal footprint, the method computes the exact Euclidean distance to the polygon boundary and assigns the sign through the inside--outside test. For a rectangle-union footprint, the min-reduced box signed distance is sign-correct: it is negative if and only if the query point lies inside at least one rectangle, zero on the exterior boundary of the union, and positive outside. For points outside the rectangle union, the exterior distance is exact. For points inside regions where multiple rectangles overlap, the interior penetration magnitude may differ from the true signed-distance magnitude, but the collision classification remains correct. In \algname{}, this signed-distance quantity is used primarily for collision screening and clearance penalization near the footprint boundary.

\begin{figure*}[t]
    \centering
    \includegraphics[width=0.9\textwidth]{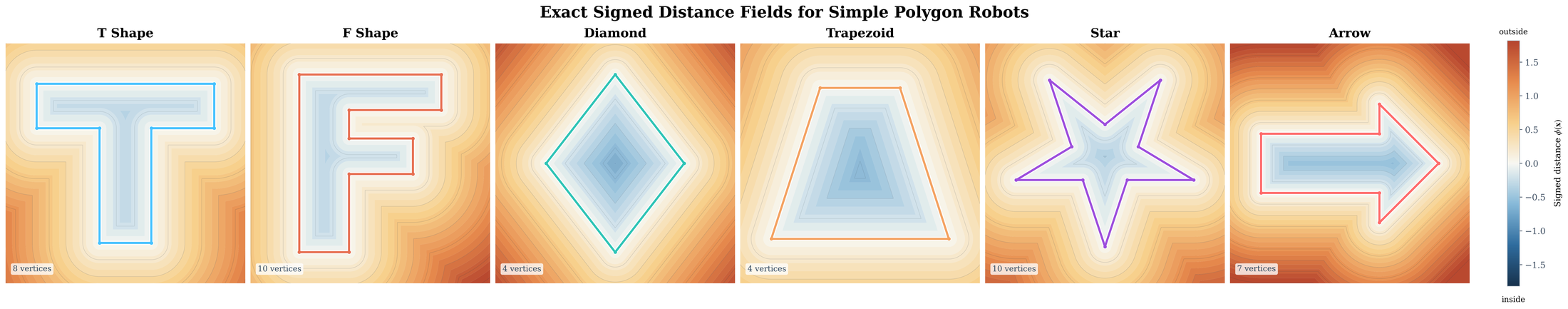}
    \caption{Representative signed-distance fields produced by the polygon-edge evaluator on six simple polygons of varying complexity. The gallery spans rectilinear (T, F), convex (Diamond, Trapezoid), and non-convex (Star, Arrow) footprints.}
    \label{fig:polygon_sdf_gallery}
\end{figure*}

Figure~\ref{fig:polygon_sdf_gallery} shows signed-distance fields produced by the polygon-edge evaluator on six footprints of varying geometric complexity. The gallery spans rectilinear (T, F), convex (Diamond, Trapezoid), and non-convex (Star, Arrow) simple polygons. A single edge-based evaluator handles all of them without convex decomposition, inflation, or shape-specific tuning. The rectilinear examples (T, F) would additionally admit the faster rectangle-cover route of Fig.~\ref{fig:t_shape_distance_methods} when computational budget is tight.

\subsection{Batched Rollout and Distance Computation}
\label{subsec:batched_computation}

The dominant computation at each control cycle is the joint evaluation of $K$ sampled rollouts, $T$ horizon steps per rollout, and $N$ obstacle points. Together these produce $K \times T \times N$ signed-distance queries. The preprocessed obstacle set is stored as a fixed-size array of $N$ points with an accompanying validity mask. Padding is used when the number of observed obstacle points is smaller than $N$, and the mask prevents padded entries from contributing to the distance reduction. The obstacle array is then broadcast over the rollout and horizon dimensions, producing $K \times T \times N$ body-frame query points $\vect{p}_{h,i}^{b,(r)}$ for signed-distance evaluation.

The rollout and distance computations are implemented in batched form. For rollout propagation, the selected motion model is evaluated over the $K$ sampled control sequences. For collision evaluation, the rollout-frame transformation of Sec.~\ref{subsec:distance_evaluation} is applied over the rollout, horizon, and obstacle-point dimensions. The transformed query points are then evaluated using either the rectangle-cover signed-distance function $d^{\pm}_{\mathrm{rect}}$ or the polygonal signed-distance function $d^{\pm}_{\mathrm{poly}}$. The validity mask is applied during the reduction over obstacle points so that padded entries do not affect the result. This produces the minimum signed-distance matrix
\begin{equation}
    d_{h}^{\min,(r)},
    \qquad
    r=1,\dots,K,\quad h=0,\dots,T-1,
\end{equation}
of size $K \times T$. These values are used for rollout scoring, feasibility screening, and trajectory validation (Sec.~\ref{subsec:mppi_safety_validation}).

The signed-distance stage evaluates a batched set of $K \times T \times N$ body-frame query points. For a rectangle-cover footprint, each query point is reduced over $R$ point-to-box distances, giving computational cost $O(KTNR)$. For a polygonal footprint, each query point requires a reduction over $B$ point-to-segment distances and an inside--outside test that also scales with $B$, giving computational cost $O(KTNB)$. The arithmetic operations inside each point-to-box or point-to-segment evaluation are constant-size vector operations and therefore affect the constant factor rather than the asymptotic scaling.

Although these computations are executed efficiently in batched form on the GPU, the total number of point-to-primitive evaluations still scales with the rollout count, horizon length, obstacle-point budget, and footprint primitive count. We implement the computation in JAX~\cite{jax2018github}. Just-in-time compilation and automatic vectorization allow the per-rollout, per-horizon-step, per-point, and per-footprint-primitive operations to be written in NumPy-like syntax and compiled into optimized computations for CPU or GPU execution. Compilation occurs at the first control cycle for a fixed input shape, and subsequent cycles reuse the compiled computation.

\subsection{Safety Penalties and Trajectory Validation}
\label{subsec:mppi_safety_validation}
The signed-distance quantities computed during rollout evaluation are incorporated into MPPI through both a soft obstacle penalty and a hard post-update validation step. For a rollout state with minimum signed distance $d$, we use the obstacle penalty
\begin{equation}
    \phi_{\mathrm{obs}}(d)
    =
    w_{\mathrm{coll}}\,\mathbb{I}(d < 0)
    +
    w_{\mathrm{rep}}\,
    \max(d_{\mathrm{safe}} - d, 0)^2,
\end{equation}
where $w_{\mathrm{coll}}$ is a large collision penalty, $w_{\mathrm{rep}}$ weights the clearance penalty, $d_{\mathrm{safe}}$ is the desired safety margin, and $\mathbb{I}(\cdot)$ denotes the indicator function ($1$ when its argument is true and $0$ otherwise). The first term penalizes penetration of the effective footprint, while the second term discourages trajectories that pass closer than the safety margin.

In addition to this soft penalty, rollout scoring records an unsafe flag $\chi^{(r)}$ whenever any horizon step violates the safety margin,
\begin{equation}
    \chi^{(r)}
    =
    \bigvee_{h=0}^{T-1}
    \left(d_{h}^{\min,(r)} < d_{\mathrm{safe}}\right).
\end{equation}
The rollout cost used for MPPI weighting is then augmented as
\begin{equation}
    \tilde{J}^{(r)}
    =
    J^{(r)}
    +
    w_{\mathrm{inf}}\,
    \mathbb{I}(\chi^{(r)}),
\end{equation}
where $w_{\mathrm{inf}}$ is chosen large enough that unsafe rollouts receive negligible importance weight.

After the nominal control sequence is updated by the MPPI weighted-average rule, the corresponding nominal trajectory is rolled out once and validated using the same clearance condition,
\begin{equation}
    d_{h}^{\min,\mathrm{nom}} \ge d_{\mathrm{safe}},
    \qquad
    h=0,\dots,T-1,
\end{equation}
where $d_{h}^{\min,\mathrm{nom}}$ denotes the minimum signed distance evaluated along the updated nominal trajectory (the rollout superscript is dropped because there is a single nominal trajectory).
If this validation fails, the controller executes a zero-velocity hold for the current cycle and reinitializes the nominal control sequence to zero for the next replanning step. If validation succeeds, the first command of the updated sequence is executed, and the remaining sequence is shifted forward in standard receding-horizon fashion. Algorithm~\ref{alg:mppi_mosaic} summarizes one control cycle of \algname{}.

\begin{algorithm}[t]
    \caption{Batched \algname{} Control Cycle}
    \label{alg:mppi_mosaic}
    \SetAlgoLined
    \textbf{Input:} pose $\vect{q}_0$, nominal $\mathbb{U}$, obstacles $\mathcal{O}$, footprint $\mathcal{B}_\mathrm{eff}$, motion model $m$\\
    \textbf{Params:} $K, T, \Delta t, \lambda, d_\mathrm{safe}, w_\mathrm{coll}, w_\mathrm{rep}, w_\mathrm{inf}$\\
    \textbf{Output:} control $\vect{u}^*$

    \tcc{1. Batched rollout propagation}
    Sample $\boldsymbol{\epsilon}_h^{(r)}$; form $\vect{u}_h^{(r)} \leftarrow \vect{u}_h + \boldsymbol{\epsilon}_h^{(r)}$; propagate $\vect{q}_{h+1}^{(r)} \leftarrow \vect{q}_h^{(r)} + \mathcal{F}_m(\vect{q}_h^{(r)},\vect{u}_h^{(r)})\,\Delta t$ from $\vect{q}_0^{(r)} = \vect{q}_0$\;

    \tcc{2. Signed-distance evaluation and scoring}
    Initialize $J^{(r)} \leftarrow 0$, $\chi^{(r)} \leftarrow \mathbf{false}$\;
    \For{all $(r, h)$}{
    Transform $\mathcal{O}$ into the body frame at $\vect{q}_h^{(r)}$ to obtain $\vect{p}_{h,i}^{b,(r)}$\;
    $d_h^{\min,(r)} \leftarrow \min_i d^\pm(\vect{p}_{h,i}^{b,(r)}, \mathcal{B}_\mathrm{eff})$\;
    $J^{(r)} \leftarrow J^{(r)} + \phi_\mathrm{task}(\vect{q}_h^{(r)},\vect{u}_h^{(r)}) + \phi_\mathrm{ctrl}(\vect{u}_h^{(r)}) + \phi_\mathrm{obs}(d_h^{\min,(r)})$\;
    $\chi^{(r)} \leftarrow \chi^{(r)} \lor (d_h^{\min,(r)} < d_\mathrm{safe})$\;
    }

    \tcc{3. Feasibility flag and path-integral update}
    $\tilde{J}^{(r)} \leftarrow J^{(r)} + w_\mathrm{inf}\,\mathbb{I}(\chi^{(r)})$;\quad $\beta \leftarrow \min_r \tilde{J}^{(r)}$\;
    $\omega^{(r)} \leftarrow \exp(-(\tilde{J}^{(r)}-\beta)/\lambda)$; normalize so $\sum_r \omega^{(r)} = 1$\;
    $\vect{u}_h \leftarrow \vect{u}_h + \sum_r \omega^{(r)}\,\boldsymbol{\epsilon}_h^{(r)}$ for $h=0,\dots,T-1$\;

    \tcc{4. Validation and execution}
    Roll out updated $\mathbb{U}$; evaluate $d_h^{\min,\mathrm{nom}}$\;
    \eIf{$d_h^{\min,\mathrm{nom}} \ge d_\mathrm{safe}\ \forall h$}{
    $\vect{u}^* \leftarrow \vect{u}_0$;\ shift $\mathbb{U}$ forward\;
    }{
    $\vect{u}^* \leftarrow \vect{0}$;\ reset $\mathbb{U} \leftarrow \vect{0}$\;
    }
    \Return $\vect{u}^*$\;
\end{algorithm}

\subsection{Extension to Hybrid-Mode Platforms}
\label{subsec:hybrid_mode}
Some mobile platforms \cite{agilex_ranger_mini} achieve near-holonomic maneuverability through a finite set of non-skidding motion modes (e.g., dual-Ackermann steering, lateral parallel motion, spin-in-place). Treating the command space as fully continuous can produce motions that require wheel slip to execute. Selecting among discrete modes avoids this but introduces mode-switching decisions. We therefore extend MPPI to a hybrid-mode formulation that evaluates each mode as a separate rollout family and selects among them at each control cycle.

Let $\mathcal{M}_{\mathrm{hyb}}$ denote the set of active motion modes. For each $m \in \mathcal{M}_{\mathrm{hyb}}$, \algname{} runs the Algorithm~\ref{alg:mppi_mosaic} cycle under dynamics $\mathcal{F}_m$, producing a validated candidate sequence $\mathbb{U}_m$ with cost $J_m$. The signed-distance evaluator and effective footprint are shared across modes; only the rollout dynamics and admissible command structure differ.

To discourage unnecessary switching, each candidate cost is augmented with a switching penalty relative to the previously active mode $m_{\mathrm{prev}}$,
\begin{equation}
    \bar{J}_m = J_m + \lambda_{\mathrm{switch}}\,\mathbb{I}{(m \neq m_{\mathrm{prev}})}.
\end{equation}
Candidates that fail trajectory validation are assigned $\bar{J}_m = +\infty$ and excluded. A cooldown variable further blocks mode changes for a fixed number of replanning steps after a switch. The selected mode is $m^\star = \arg\min_m \bar{J}_m$.

For physical deployment, the chosen command is projected onto the command structure of mode $m^\star$ and passed through a deadzone correction. If its magnitude falls below a mode-dependent threshold, it is scaled to the minimum executable value. The direction is preserved for translational modes and the sign is preserved for pure rotation. This post-processing affects only hardware execution, not the rollout-time signed-distance evaluation. Algorithm~\ref{alg:hybrid_mppi} summarizes the full procedure.

\begin{algorithm}[t]
    \caption{Hybrid-Mode \algname{} with Actuator Post-Processing}
    \label{alg:hybrid_mppi}
    \SetAlgoLined
    \textbf{Input:} pose $\vect{q}_0$, sequences $\{\mathbb{U}_m\}_{m \in \mathcal{M}_\mathrm{hyb}}$, obstacles $\mathcal{O}$, footprint $\mathcal{B}_\mathrm{eff}$, $m_\mathrm{prev}$, $\tau_\mathrm{cool}$\\
    \textbf{Params:} $\lambda_\mathrm{switch}, \tau_\mathrm{cool}^{\max}, v_{\min}, \omega_{\min}, \delta_\mathrm{noise}^v, \delta_\mathrm{noise}^\omega$ (plus Algorithm~\ref{alg:mppi_mosaic} parameters)\\
    \textbf{Output:} command $\vect{u}^*$, mode $m^*$

    \tcc{1. Per-mode rollout evaluation}
    \ForEach{$m \in \mathcal{M}_\mathrm{hyb}$}{
        Run Algorithm~\ref{alg:mppi_mosaic} with dynamics $\mathcal{F}_m$ to obtain $\mathbb{U}_m, J_m, \nu_m$\;
        $\bar{J}_m \leftarrow J_m + \lambda_\mathrm{switch}\,\mathbb{I}(m \neq m_\mathrm{prev})$ if $\nu_m$, else $+\infty$\;
    }

    \tcc{2. Mode selection with cooldown}
    \lIf{$\bar{J}_m = +\infty\ \forall m$}{\Return $\vect{0},\, m_\mathrm{prev}$ \tcp*{safe stop}}
    $m^* \leftarrow \arg\min_m \bar{J}_m$\;
    \lIf{$\tau_\mathrm{cool} > 0$ and $m^* \neq m_\mathrm{prev}$}{$m^* \leftarrow m_\mathrm{prev}$}
    $\tau_\mathrm{cool} \leftarrow \tau_\mathrm{cool}^{\max}$ if $m^* \neq m_\mathrm{prev}$, else $\max(\tau_\mathrm{cool}-1, 0)$\;
    $\vect{u}_\mathrm{raw} \leftarrow$ first command of $\mathbb{U}_{m^*}$\;

    \tcc{3. Actuator post-processing}
    $\vect{u}^* \leftarrow \textsc{ProjectToMode}(\vect{u}_\mathrm{raw}, m^*)$ \tcp*{drop components inadmissible for $m^*$}
    Let $\vect{u}_\mathrm{lin}^*, \omega^*$ denote the linear and angular parts of $\vect{u}^*$\;
    \eIf{$m^*$ is translational}{
        $v_\mathrm{mag} \leftarrow \|\vect{u}_\mathrm{lin}^*\|_2$;\quad
        \lIf{$\delta_\mathrm{noise}^v < v_\mathrm{mag} < v_{\min}$}{$\vect{u}_\mathrm{lin}^* \leftarrow (v_{\min}/v_\mathrm{mag})\,\vect{u}_\mathrm{lin}^*$}
    }{
        \lIf{$\delta_\mathrm{noise}^\omega < |\omega^*| < \omega_{\min}$}{$\omega^* \leftarrow \operatorname{sgn}(\omega^*)\,\omega_{\min}$}
    }
    \Return $\vect{u}^*, m^*$\;
\end{algorithm}

\section{Experiments and Results}
\label{sec:experiments_and_results}

\subsection{Experimental Design}

We evaluate \algname{} in simulation and real-world settings across multiple robot platforms with distinct footprint geometries and kinematic characteristics. The experiments are organized around four questions:

\begin{enumerate}
    \item \textit{Computational efficiency:} Can the JAX-based batched signed-distance evaluator support real-time MPPI rollout evaluation under a fixed sampling budget?

    \item \textit{Footprint-modeling fidelity:} Does explicit footprint-aware signed-distance evaluation preserve feasible motion in clearance-limited environments where convex-hull or simplified footprint approximations become overly conservative?

    \item \textit{Cross-platform deployment:} Can the same collision-evaluation and MPPI update structure be reused across platforms by changing only the explicit footprint representation and the platform-specific rollout model, including cases with carried objects, payloads, or implements?

    \item \textit{Hybrid-motion adaptability:} Can the same footprint-aware MPPI formulation be extended to platforms with multiple kinematic modes, enabling mode selection through rollout-cost evaluation rather than manually designed mode-specific navigation rules?
\end{enumerate}

Dynamic-obstacle experiments are included as a supporting evaluation. Since \algname{} does not explicitly predict obstacle motion, moving obstacles are treated as quasi-static within each MPPI horizon. Reactivity is obtained through receding-horizon replanning from updated point-cloud observations. Therefore, the dynamic-obstacle results should be interpreted within the low-speed settings considered in this work.

\textbf{Experimental Platforms and Environments.}
We conduct experiments in simulation and on real robot platforms to evaluate \algname{} under different footprint geometries, motion models, sensing conditions, and obstacle configurations. In simulation, controlled narrow-passage scenarios are used to evaluate footprint-modeling fidelity under different degrees of narrowness, while dynamic-obstacle scenarios are used to evaluate reactive replanning from updated point-cloud observations. IR-SIM~\cite{han2025neupan} is used as the main lightweight simulator, and Gazebo~\cite{koenig2004design} is used for the higher-fidelity dynamic-obstacle comparison. To support the omni-directional body-velocity cases considered in this work, we adapt IR-SIM with a control interface that accepts longitudinal, lateral, and yaw-rate commands.

\begin{table*}[t]
    \caption{Cross-platform motion models and kinematic limits used in the hardware deployment experiments.}
    \label{tab:platform_kinematic_setup}
    \centering
    \small
    \setlength{\tabcolsep}{4pt}
    \renewcommand{\arraystretch}{1.08}
    \begin{adjustbox}{max width=\textwidth}
        \begin{tabular}{lccccc}
            \toprule
            \multirow{2}{*}{Feature}               & \multirow{2}{*}{Dual-arm robot} & \multicolumn{3}{c}{AgileX Ranger mini} & \multirow{2}{*}{Unitree Go2}                                                         \\
                                                   &                                 & Ack.                                   & Parallel                     & Spin                      &                           \\
            \midrule
            Motion model(s)                        & Differential mode               & Dual-Ackermann mode                    & Parallel mode                & Spin mode                 & Omni-motion mode          \\
            Control input(s)                       & $(v_x,\omega)$                  & $(v_x,\omega)$                         & $(v_x,v_y)$                  & $(\omega)$                & $(v_x,v_y,\omega)$        \\
            Longitudinal velocity limit (m/s)      & $v_x \in [-1.5,1.5]$            & $v_x \in [-1.5,1.5]$                   & $v_x \in [-1.0,1.0]$         & $v_x = 0$                 & $v_x \in [-1.0,1.0]$      \\
            Lateral velocity limit (m/s)           & Not used                        & Not used                               & $v_y \in [-0.6,0.6]$         & Not used                  & $v_y \in [-0.4,0.4]$      \\
            Yaw-rate limit (rad/s)                 & $\omega \in [-1.0,1.0]$         & $\omega \in [-1.0,1.0]$                & $\omega = 0$                 & $\omega \in [-1.0,1.0]$   & $\omega \in [-1.0,1.0]$   \\
            Linear acceleration limit (m/s$^2$)    & $a_x \in [-1.0,1.0]$            & $a_x \in [-1.0,1.0]$                   & $a_x,a_y \in [-1.0,1.0]$     & Not used                  & $a_x,a_y \in [-1.0,1.0]$  \\
            Angular acceleration limit (rad/s$^2$) & $a_\omega \in [-1.0,1.0]$       & $a_\omega \in [-1.0,1.0]$              & Not used                     & $a_\omega \in [-2.0,2.0]$ & $a_\omega \in [-1.0,1.0]$ \\
            \bottomrule
        \end{tabular}
    \end{adjustbox}
\end{table*}

\begin{figure}[t]
    \centering
    \includegraphics[width=1.0\columnwidth]{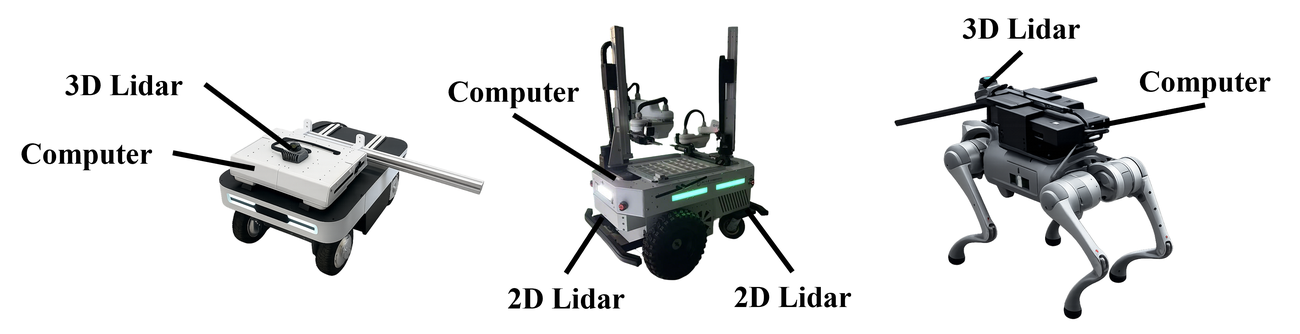}
    \caption{The three mobile robot platforms used in the hardware experiments: AgileX Ranger mini, differential-drive dual-arm robot, and Unitree Go2 quadrupedal robot.}
    \label{fig:experiment_platform}
\end{figure}

The hardware experiments use three platforms with distinct motion interfaces and effective footprints, as shown in Fig.~\ref{fig:experiment_platform}. The differential-drive dual-arm robot represents a conventional indoor service platform. AgileX Ranger mini is a hybrid 4WS/4WD platform with dual-Ackermann, parallel, and spin-in-place modes, and is used to evaluate the hybrid-mode extension of \algname{}. The Unitree Go2 quadruped carries a rigid bar that extends its projected footprint, providing a body-velocity deployment case with a task-dependent footprint. Across these platforms, the same collision-evaluation principle is used: observed obstacle points are transformed into the predicted robot body frame and evaluated against an explicit planar effective footprint represented by either a rectangle cover or a simple polygon.

\textbf{Shared MPPI Sampling Budget.}
Unless otherwise stated, the navigation experiments use $K=1000$ sampled trajectories, $T=50$ horizon steps, and $\Delta t=0.1$~s, corresponding to a 5~s prediction horizon. The local point cloud is preprocessed and downsampled to $N=100$ obstacle points before rollout evaluation. Thus, each control cycle evaluates $KTN=5.0 \times 10^6$ point-to-footprint signed-distance queries. In our implementation, this setting occupies approximately 500~MB of GPU memory.

All platforms are equipped with onboard 2D or 3D LiDAR sensors for local point-cloud acquisition. The weak guidance path is provided to the local planner and transformed into the robot coordinate frame using the available state-estimation or SLAM system, such as Cartographer~\cite{hess2016real} or FAST-LIO2~\cite{xu2022fast}. The guidance provides high-level directional information, while local collision avoidance and maneuver generation are handled by \algname{}.

\textbf{Evaluation Metrics.}
We report four task-performance metrics and one clearance-difficulty metric. Task-specific measures, such as path length, are reported only when they are needed to interpret a particular experiment.

\begin{itemize}
    \item \textbf{Success rate} is the ratio of successful trials to the total number of trials. A trial is counted as successful if the robot reaches the target pose within the prescribed position and heading tolerances, remains collision-free with respect to the observed obstacle set, and completes the task within the predefined time limit. Collision, timeout, or failure to make progress is counted as failure.

    \item \textbf{Navigation time} is the elapsed time required to complete the navigation task successfully. In simulation, it is measured from the control-loop steps; in real-world experiments, it is recorded from system timestamps.

    \item \textbf{Mean speed} is the average speed along the executed trajectory, computed over successful trials.

    \item \textbf{Path length} is the total executed trajectory length. It is reported for experiments where route efficiency is compared in addition to completion time.

    \item \textbf{Degree of Narrowness (DoN)}
          \label{subsec:don_metric}
          quantifies the difficulty of clearance-limited navigation. Following~\cite{han2025neupan}, we define
          \begin{equation}
              \mathrm{DoN} = W_r / W_p,
          \end{equation}
          where $W_r$ is the effective robot width and $W_p$ is the minimum passable width of the environment. Larger DoN values indicate tighter passages; as $\mathrm{DoN}\rightarrow 1$, the available clearance becomes small and footprint modeling becomes more important.

          The effective width depends on the motion model because it is measured relative to the direction of translation. Let $\mathbf{d}$ denote the translation direction and $\mathbf{n}$ an orthogonal unit vector. For a footprint $S \subset \mathbb{R}^2$, the directional width is
          \begin{equation}
              W_r(\mathbf{n}) =
              \max_{\mathbf{x}\in S} \mathbf{n}^{\top}\mathbf{x}
              -
              \min_{\mathbf{x}\in S} \mathbf{n}^{\top}\mathbf{x},
          \end{equation}
          and $W_p$ is measured along the same cross-sectional direction. For Ackermann-steered and differential-drive robots, $\mathbf{d}$ is the forward direction, so $W_r$ is the lateral body width. For parallel motion, $\mathbf{d}$ is the sideways direction, so $W_r$ is the longitudinal span. For omni-directional motion, we report the minimum directional width over all planar translation directions.

          For concave footprints, this directional width is a coarse scalar measure rather than a complete description of passability. In particular, $W_r(\mathbf{n})$ is unchanged if $S$ is replaced by its convex hull, since both have the same extrema along direction $\mathbf{n}$. Therefore, DoN is used only to quantify passage tightness; it does not capture the extra feasible configurations that may be available when the planner reasons about the exact concave footprint instead of its convex hull.
\end{itemize}

\subsection{Experiment 1: Benchmark the Signed-Distance Evaluator}

This experiment evaluates the signed-distance evaluator, which is the dominant geometric operation in the proposed MPPI rollout computation. The goal is to assess runtime, scaling behavior, and deployment overhead rather than distance accuracy, since the analytic distance formulation is described in Sec.~\ref{subsec:distance_evaluation}.

We compare \algname{} with the deep unfolded neural encoder (DUNE) from NeuPAN~\cite{han2025neupan}, a recent learning-based method for point-to-robot distance computation. Both methods take obstacle points and robot geometry as inputs and return point-to-robot distance values without constructing an occupancy grid or ESDF. The comparison therefore isolates two different design choices for the distance module: analytic evaluation from an explicit footprint representation versus learned distance approximation.

The experiment addresses three questions:
\begin{enumerate}
    \item[(Q1)] How does analytic signed-distance evaluation compare with DUNE in runtime and scaling behavior?
    \item[(Q2)] How much does the rectangle-cover specialization accelerate evaluation on rectilinear footprints compared with direct polygon-edge evaluation?
    \item[(Q3)] What deployment overhead is required when switching footprints or robot platforms?
\end{enumerate}

\textbf{Benchmark setup.}
All measurements are collected on an Ubuntu workstation with an Intel Core i7-13700F CPU and an NVIDIA GeForce RTX 4060 Ti GPU. For each benchmarked footprint, query points are uniformly sampled from a $50 \times 50$~m region. Unless otherwise stated, per-call inference time is averaged over 50 randomly sampled batches, and \algname{} and DUNE use identical query batches for fairness. The DUNE models are trained using the default training settings reported for NeuPAN~\cite{han2025neupan}. For Q1, we use four representative footprints: a rectangle, a trapezoid, a sprayer footprint, and a double-sided pruner footprint. For Q2, we use three rectilinear multi-component footprints: L-, T-, and F-shaped bodies.

\subsubsection{Analytic versus Learned Distance Evaluation (Q1)}

\textbf{Property comparison.}
Table~\ref{tab:distance_feature_compare} summarizes the structural differences between the two evaluators. DUNE uses a learned surrogate for distance evaluation and requires footprint-specific training or adaptation. In contrast, \algname{} evaluates distances analytically from an explicit footprint representation. For simple-polygon footprints, the signed distance is computed from point-to-edge distances and an inside--outside test. For rectangle-cover footprints, the exterior distance is exact with respect to the rectangle union, while the interior value is used primarily for collision classification and penalty assignment.

\begin{table}[t]
    \caption{Property comparison of DUNE and the proposed distance evaluator.}
    \label{tab:distance_feature_compare}
    \centering
    \setlength{\tabcolsep}{4pt}
    \renewcommand{\arraystretch}{1.05}
    \resizebox{\columnwidth}{!}{%
        \begin{tabular}{lcc}
            \toprule
            Property                  & DUNE~\cite{han2025neupan} & \algname{}                   \\
            \midrule
            Pre-training required     & Yes                       & No                           \\
            Distance evaluator        & Learned surrogate         & Analytic geometry            \\
            Distance representation   & Approximate               & Explicit footprint-based     \\
            Convex footprint support  & Yes                       & Yes                          \\
            Concave footprint support & No                        & Yes                          \\
            Footprint change          & Retraining                & Update footprint description \\
            \bottomrule
        \end{tabular}%
    }
\end{table}

\textbf{Runtime comparison.}
At the representative benchmark size of 100{,}000 query points, the JAX-batched analytic evaluator is consistently faster than DUNE on the GPU. The measured speedups are $14.0\times$ for the rectangle, $12.6\times$ for the trapezoid, $18.9\times$ for the sprayer footprint, and $16.0\times$ for the double-sided pruner footprint.

\begin{figure}[t]
    \centering
    \includegraphics[width=0.98\columnwidth]{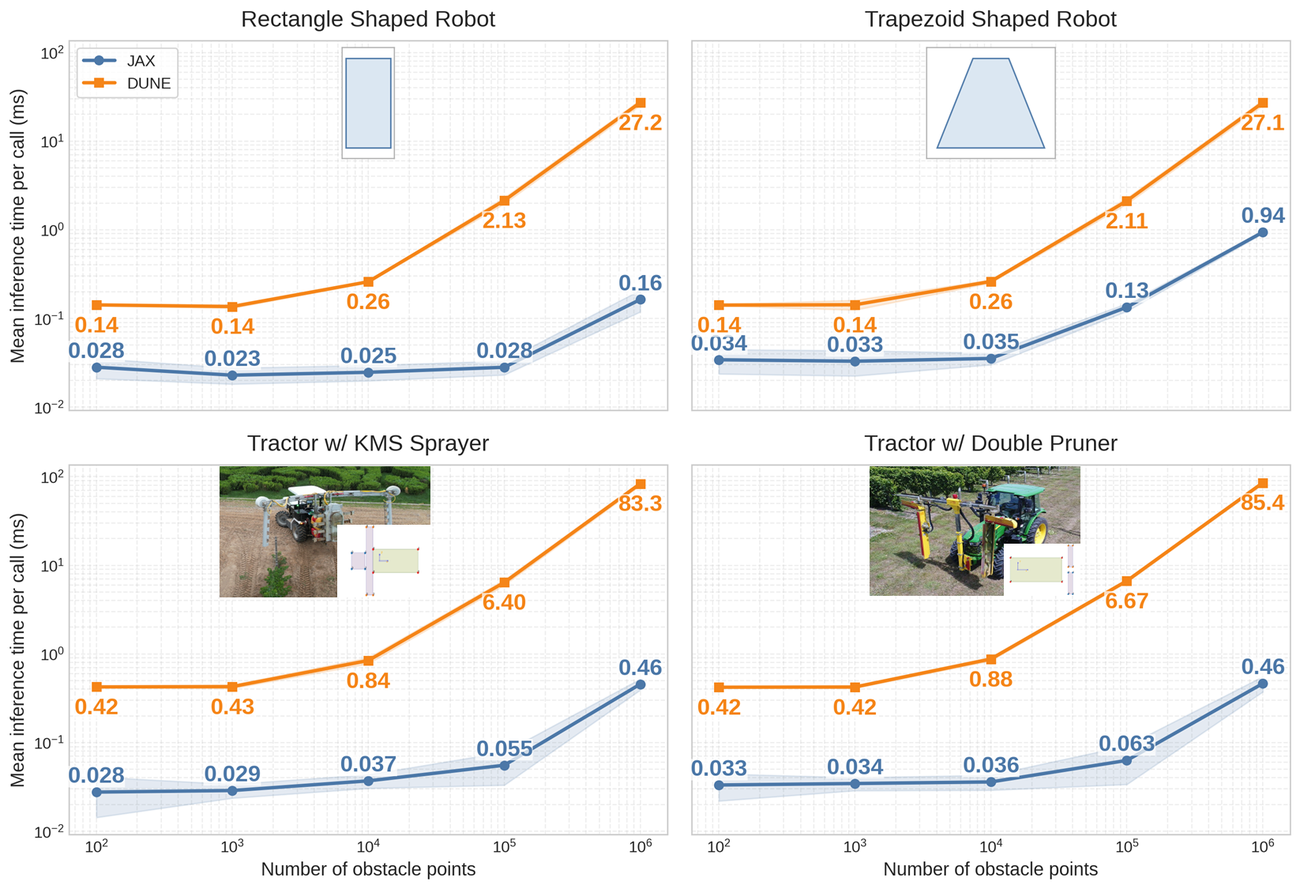}
    \caption{Inference time versus obstacle point count for DUNE and the JAX-based evaluator across four representative footprints.}
    \label{fig:curve_comparison_to_dune}
\end{figure}

\textbf{Scaling with obstacle point count.}
Figure~\ref{fig:curve_comparison_to_dune} reports per-call inference time as the number of query points grows from $10^2$ to $10^6$. The curves show the mean over 100 trials, with shaded bands denoting one standard deviation. Across the tested range, \algname{} scales more favorably than DUNE for both simple and multi-part footprints. This behavior is expected because the proposed evaluator uses explicit analytic geometry implemented as tensorized arithmetic and reduction operations, whereas DUNE performs inference through a neural encoder. As the query batch size increases, the neural inference overhead becomes more pronounced, while the analytic evaluator remains well matched to GPU-parallel batched distance computation.

\subsubsection{Rectangle-Cover versus Polygon Evaluation}

For rectilinear footprints, \algname{} provides a rectangle-cover specialization in addition to the general simple-polygon evaluator. Table~\ref{tab:rect_poly_gpu_compare} reports per-call GPU runtime on the tested L-, T-, and F-shaped footprints. Under the same benchmark setup, the rectangle-cover path is faster than direct polygon-edge evaluation by $3.25\times$, $3.34\times$, and $2.03\times$, respectively. These results justify using rectangle covers as a computational specialization for rectilinear bodies, while retaining the polygon-edge route for general planar polygonal footprints.

\begin{table}[t]
    \caption{GPU runtime comparison of rectangle-cover and polygon-edge signed-distance evaluation on rectilinear footprints.}
    \label{tab:rect_poly_gpu_compare}
    \centering
    \resizebox{\columnwidth}{!}{%
        \begin{tabular}{lccc}
            \toprule
            Shape   & Rectangle cover    & Polygon edges      & Speedup      \\
            \midrule
            L-shape & $0.09 \pm 0.11$ ms & $0.29 \pm 0.23$ ms & 3.25$\times$ \\
            T-shape & $0.09 \pm 0.12$ ms & $0.32 \pm 0.20$ ms & 3.34$\times$ \\
            F-shape & $0.15 \pm 0.12$ ms & $0.30 \pm 0.11$ ms & 2.03$\times$ \\
            \bottomrule
        \end{tabular}%
    }
\end{table}

\subsubsection{Deployment Overhead}

The proposed evaluator requires no neural-network training and no precomputed distance field. A new platform requires specifying the footprint representation and motion model, followed by JAX just-in-time compilation at the first execution for the fixed input shape. In our setup, this compilation takes less than 1~s, and subsequent calls reuse the compiled computation. By contrast, DUNE requires a trained distance encoder for the target footprint representation. Under the default settings reported in~\cite{han2025neupan}, training takes approximately one hour per footprint. Therefore, when the robot footprint changes substantially, \algname{} requires only updating the explicit footprint description, whereas the learned distance module requires retraining or adaptation.

Experiment~1 shows that the analytic evaluator provides order-of-magnitude GPU speedups over DUNE on the tested footprints, scales consistently with the number of query points, and supports fast deployment by avoiding footprint-specific training. The rectangle-cover specialization further accelerates evaluation on rectilinear footprints by $2.03$--$3.34\times$ compared with direct polygon-edge evaluation. These results support the use of analytic signed-distance evaluation as a lightweight, training-free distance module for MPPI-based local navigation.

\subsection{Experiment 2: Clearance-Limited Navigation with Exact versus Convex Footprints}

This experiment evaluates the footprint-modeling claim in clearance-limited navigation. The first two test cases isolate the effect of explicit footprint modeling in narrow passages, while the last two test cases evaluate the same principle under mixed static-dynamic obstacle settings. The full setup includes four simulation studies: a differential-drive corridor DoN sweep, an omni-directional narrow-gap benchmark, an IR-SIM dynamic-obstacle corridor benchmark, and a Gazebo dynamic-obstacle comparison.

\subsubsection{Test Case 1: Differential-Drive Corridor with DoN Sweeps}

A differential-drive robot with a T-shaped footprint navigates through a cluttered corridor with varying degrees of narrowness. We create synthetic environments with gap widths tuned to achieve DoN values from 0.6 to 1.0, while keeping the start pose, goal pose, kinematic limits, and overall layout fixed. We compare three methods: (i) \algname{} with the T-shaped footprint represented explicitly, (ii) Convex-MPPI with the convex hull of the T-shaped footprint, and (iii) NeuPAN with the same convex-hull footprint representation.

For NeuPAN, the DUNE distance encoder is trained for the convex-hull T-shaped footprint. The Learnable Optimization Network (LON) is used to tune the NRMP planner parameters on a representative base case, DoN~=~0.9. The tuned NRMP parameters are then kept fixed when evaluating the remaining DoN values. This protocol reflects a deployment-transfer setting in which the footprint-specific distance encoder and planner parameters are prepared once and applied across related environment variations without per-case planner retuning. NeuPAN uses a reference speed of $1.5$~m/s, chosen to be compatible with its typical update frequency of approximately 15~Hz in our setup. For the MPPI-based methods, we use the same linear-velocity limit of $\pm 2$~m/s, but do not include a fixed reference-speed critic; their executed speed is determined by rollout costs and local clearance.

For DoN~=~0.6--0.9, all three methods use the same straight-line weak guidance path, and the passage remains feasible under both the explicit T-shaped footprint and its convex hull. For DoN~=~1.0, the convex-hull representation removes the remaining feasible passage. We therefore also test this case using an $A^*$ waypoint path as weak guidance. The $A^*$ path does not account for the detailed T-shaped footprint geometry. With this guidance, \algname{} completes the task by exploiting the non-convex free space of the T-shaped footprint, while Convex-MPPI and NeuPAN still fail under their convex-hull footprint representation.

\begin{figure}[t]
    \centering
    \includegraphics[width=0.9\columnwidth]{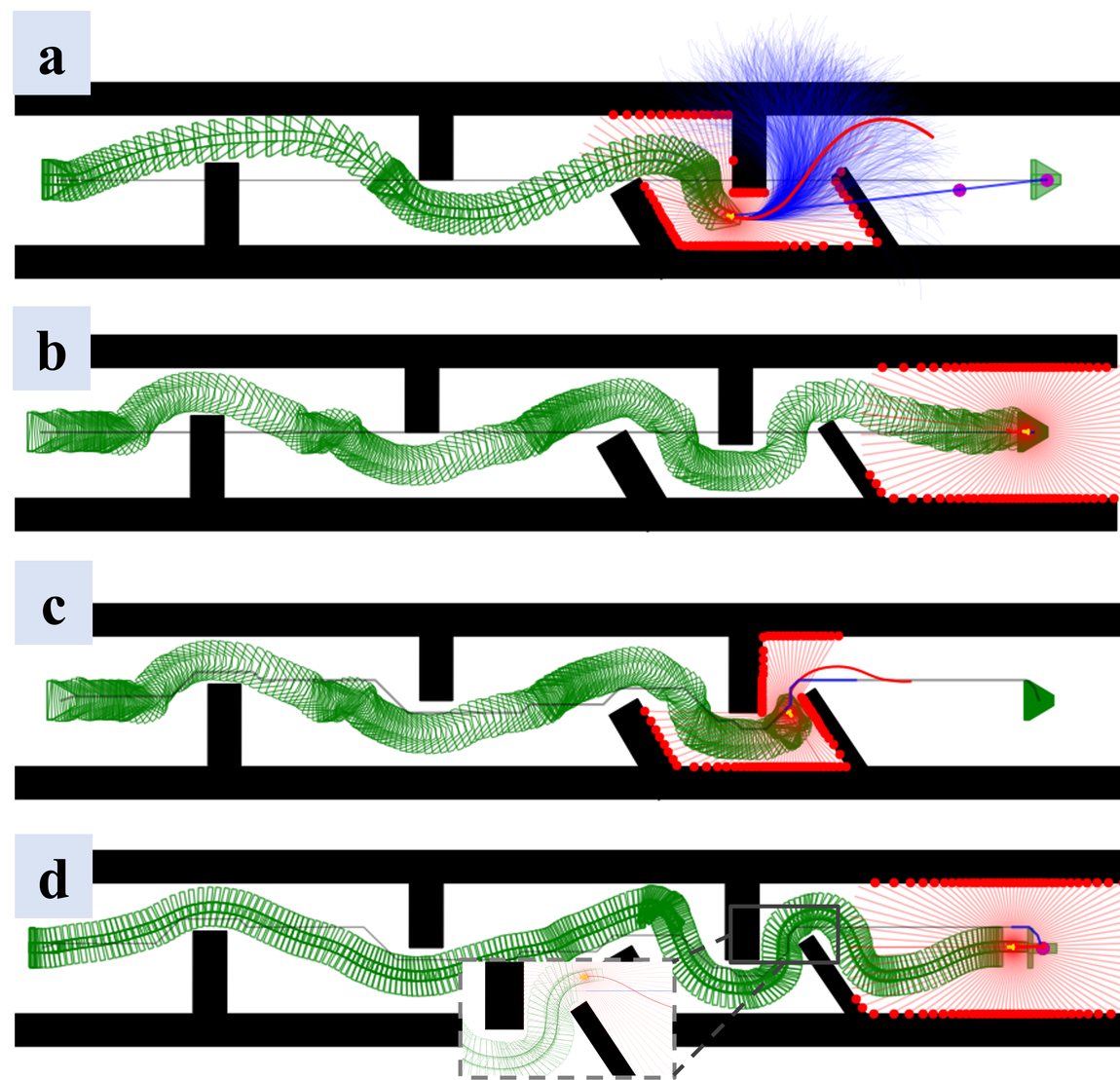}
    \caption{Representative cases from the corridor DoN sweep: (a) NeuPAN at DoN~=~0.6, (b) Convex-MPPI at DoN~=~0.6, (c) NeuPAN at DoN~=~1.0, and (d) \algname{} at DoN~=~1.0. The zoomed inset in (d) shows how the explicit T-shaped footprint preserves a narrow feasible passage that is removed by the convex-hull approximation.}
    \label{fig:anyshape_comparison}
\end{figure}

Figure~\ref{fig:anyshape_comparison} shows representative cases, and Table~\ref{tab:corridor_don_sweep} summarizes the mean speed and navigation time over successful trials. For DoN~=~0.6--0.9, all methods complete the task, and the fastest method varies across DoN levels. At DoN~=~1.0, only \algname{} succeeds, completing the task in $70.4$~s. The zoomed inset in Fig.~\ref{fig:anyshape_comparison}(d) shows how the explicit T-shaped footprint preserves a narrow feasible passage that is removed by the convex-hull approximation.

\begin{table}[t]
    \caption{Corridor DoN sweep: average speed and navigation time over successful trials. For DoN~=~0.6--0.9, all methods use straight-line weak guidance. For DoN~=~1.0, an $A^*$ waypoint path is used as weak guidance; Convex-MPPI and NeuPAN still fail under the convex-hull footprint representation.}
    \label{tab:corridor_don_sweep}
    \centering
    \small
    \setlength{\tabcolsep}{3pt}
    \renewcommand{\arraystretch}{1.1}
    \resizebox{\columnwidth}{!}{%
        \begin{tabular}{lccccccc}
            \toprule
            \multirow{2}{*}{DoN} & \multicolumn{2}{c}{\algname{}} & \multicolumn{2}{c}{Convex-MPPI} & \multicolumn{2}{c}{NeuPAN}                                          \\
            \cmidrule(lr){2-3} \cmidrule(lr){4-5} \cmidrule(lr){6-7}
                                 & Speed (m/s)                    & Time (s)                        & Speed (m/s)                & Time (s)      & Speed (m/s) & Time (s) \\
            \midrule
            0.6                  & 1.48                           & 46.5                            & \textbf{1.49}              & \textbf{46.2} & 1.03        & 65.0     \\
            0.7                  & \textbf{1.40}                  & \textbf{47.8}                   & 1.32                       & 53.0          & 1.02        & 65.8     \\
            0.8                  & 1.20                           & 57.5                            & \textbf{1.28}              & \textbf{54.1} & 1.01        & 66.2     \\
            0.9                  & 1.03                           & 67.8                            & \textbf{1.15}              & \textbf{61.1} & 0.98        & 68.3     \\
            1.0                  & \textbf{1.03}                  & \textbf{70.4}                   & Fail                       & Fail          & Fail        & Fail     \\
            \bottomrule
        \end{tabular}%
    }
\end{table}

\subsubsection{Test Case 2: Omni-Directional Gap Scenario}

An omni-directional robot with an L-shaped footprint navigates through a narrow-gap environment. The L-shaped footprint has an overall height of 2.0~m, an overall width of 2.0~m, and a uniform leg thickness of 0.4~m. The gap width varies over $[1.9,2.0,2.2,2.4]$~m, producing different DoN values while keeping the start pose, goal pose, and environment layout fixed. Omni-directional motion allows the robot to rotate and translate laterally, so it can align favorable cross-sectional dimensions with the passage constraint.

We compare two methods under the same omni-motion dynamics and MPPI settings: (i) \algname{} with the L-shaped footprint represented explicitly and (ii) Convex-MPPI with the convex hull of the L-shaped footprint. NeuPAN is not included in this test because the purpose is to isolate footprint modeling under identical omni-directional MPPI dynamics.

\begin{figure}[t]
    \centering
    \begin{subfigure}[b]{0.49\columnwidth}
        \centering
        \includegraphics[width=\linewidth]{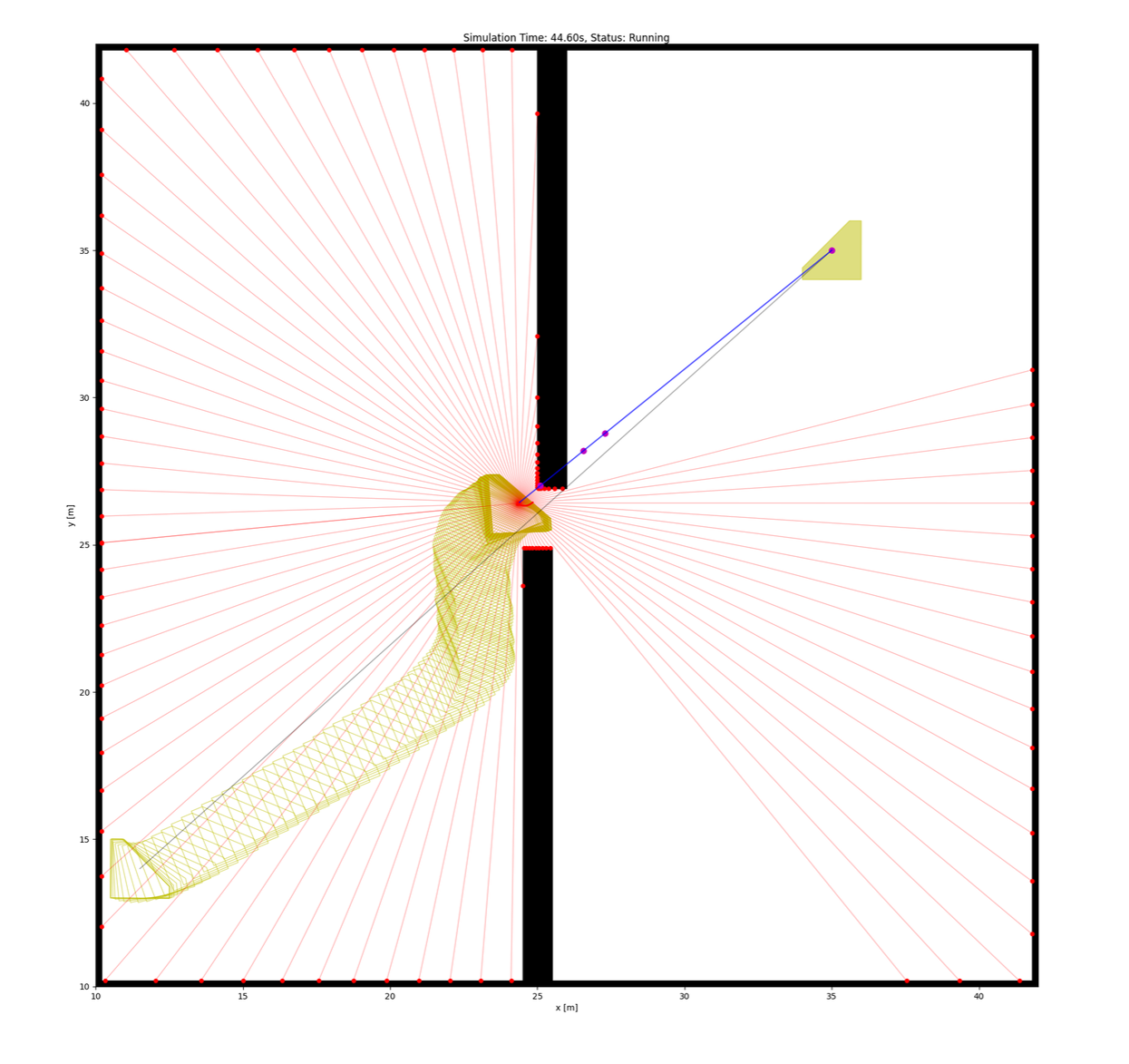}
        \caption{Convex-MPPI}
        \label{fig:gap_convex}
    \end{subfigure}\hfill%
    \begin{subfigure}[b]{0.49\columnwidth}
        \centering
        \includegraphics[width=\linewidth]{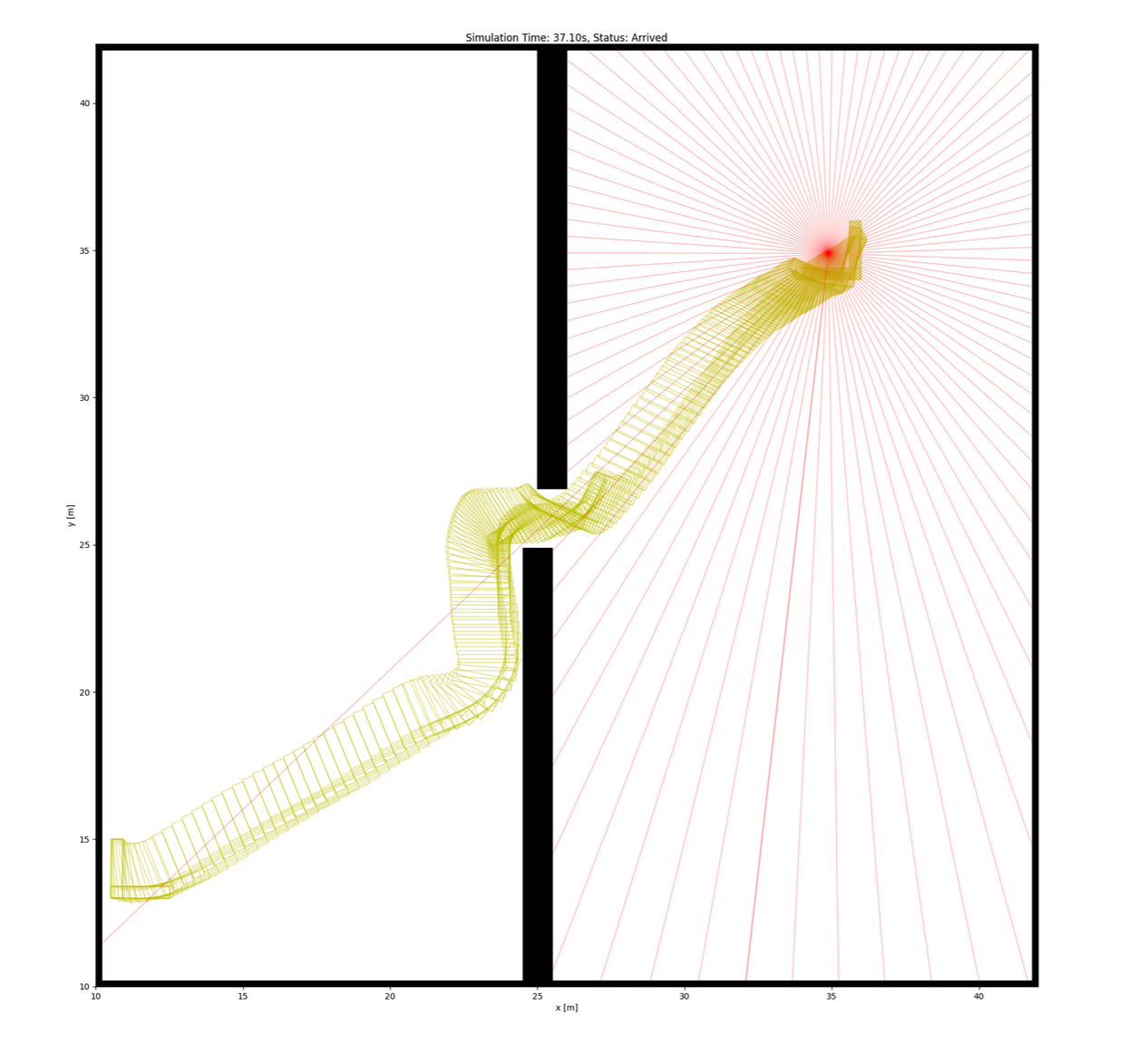}
        \caption{EXACT-MPPI}
        \label{fig:gap_mosaic}
    \end{subfigure}
    \caption{Omni-directional gap scenario in the simulator at DoN~=~1.05.}
    \label{fig:gap_scenario}
\end{figure}

Figure~\ref{fig:gap_scenario} shows the omni-directional gap scenario at DoN~=~1.05. Convex-MPPI fails because the convex-hull approximation removes the feasible passage, whereas \algname{} succeeds by evaluating the explicit L-shaped footprint. This case isolates the benefit of footprint-aware collision evaluation when the motion model is held fixed.

Table~\ref{tab:gap_don_sweep} summarizes the omni-directional gap scenario across DoN levels. When both methods are feasible, \algname{} completes the task slightly faster, reducing navigation time by about $5\%$ across the shared feasible cases. At the hardest setting, DoN~=~1.05, \algname{} still succeeds with an average speed of $0.98$~m/s and a navigation time of $20.04$~s, whereas Convex-MPPI fails. These results show that explicit footprint modeling can improve efficiency in moderately tight gaps and extend the feasible operating range when the convex-hull approximation becomes too conservative.

\begin{table}[t]
    \caption{Omni-directional gap DoN sweep comparing \algname{} and Convex-MPPI. Reported values are average speed and navigation time over successful trials; ``-'' indicates failure to complete the task at that DoN.}
    \label{tab:gap_don_sweep}
    \centering
    \small
    \renewcommand{\arraystretch}{1.1}
    \resizebox{\columnwidth}{!}{%
        \begin{tabular}{lcccc}
            \toprule
            \multirow{2}{*}{DoN} & \multicolumn{2}{c}{\algname{}} & \multicolumn{2}{c}{Convex-MPPI}                               \\
            \cmidrule(lr){2-3} \cmidrule(lr){4-5}
                                 & Speed (m/s)                    & Nav. Time (s)                   & Speed (m/s) & Nav. Time (s) \\
            \midrule
            0.83                 & 1.16                           & 17.42                           & 1.17        & 18.42         \\
            0.91                 & 1.16                           & 17.51                           & 1.16        & 18.51         \\
            1.00                 & 1.11                           & 18.18                           & 1.10        & 19.16         \\
            1.05                 & 0.98                           & 20.04                           & Fail        & Fail          \\
            \bottomrule
        \end{tabular}%
    }
\end{table}

The first two test cases support the core footprint-modeling claim. In the differential-drive corridor sweep, \algname{} is not uniformly fastest over the shared feasible DoN range: Convex-MPPI is slightly faster at several easier or moderately narrow settings, while \algname{} gives the shortest time at DoN~=~0.7 and is the only method that completes the DoN~=~1.0 case. In the omni-directional gap scenario, \algname{} reduces navigation time in the shared feasible cases and succeeds at the hardest setting where Convex-MPPI fails. The advantage is therefore most pronounced near the feasibility boundary, where convex-hull approximations can remove narrow but valid configurations.

\subsubsection{Test Case 3: Dynamic-Obstacle Corridor Benchmark}

Dynamic testing is included as a supporting study. We evaluate four navigation setups in an 8~m-wide corridor containing mixed static and dynamic obstacles: \algname{} with the explicit T-shaped footprint, Convex-MPPI with the convex hull of the T-shaped footprint, Rectangle-MPPI with a rectangle-cover approximation of the T-shaped footprint, and NeuPAN with the convex-hull T-shaped footprint. Each method is tested for 50 trials under the same environment configuration, with 2 dynamic obstacles, 8 static obstacles, and concave polygonal obstacle geometry. For guidance, we use a straight-line reference from start to goal. A run is counted as successful only if the robot reaches the goal within a 60~s time limit. Mean navigation time, path length, and speed are reported over successful runs.

For NeuPAN, the DUNE encoder uses the same convex-hull T-shaped footprint as in Test Case~1, and the NRMP parameters are kept from the LON-tuned DoN~=~0.9 corridor setting. We do not retune the NRMP parameters for each random obstacle realization. This follows the same deployment-transfer protocol as Test Case~1: the baseline planner is configured on a representative case and then evaluated under related environment variations.

\begin{figure}[t]
    \centering
    \includegraphics[width=0.85\columnwidth]{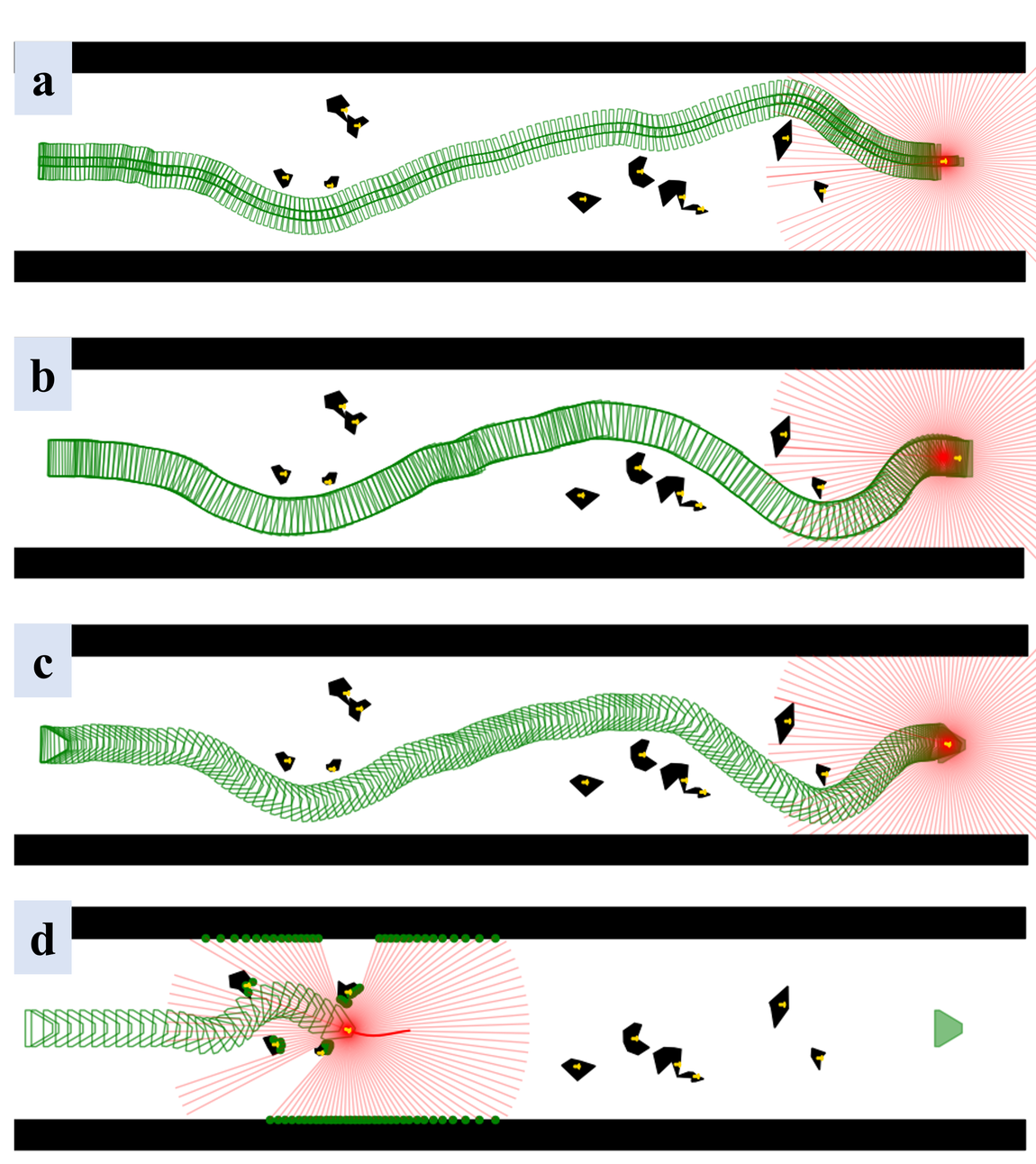}
    \caption{IR-SIM comparison in the 8~m-wide corridor with mixed static and dynamic obstacles: (a) \algname{}, (b) Convex-MPPI, (c) Rectangle-MPPI, and (d) NeuPAN with a convex-hull T-shaped footprint.}
    \label{fig:irsim_dynamic_geometry_compare}
\end{figure}

Figure~\ref{fig:irsim_dynamic_geometry_compare} shows representative qualitative behavior in this benchmark. Quantitative results are summarized in Table~\ref{tab:sim_anyshape_compare}. \algname{} achieves the highest success rate, $0.92$, with mean navigation time $44.25$~s, mean path length $63.62$~m, and mean speed $1.482$~m/s. Convex-MPPI reaches success rate $0.86$; Rectangle-MPPI reaches success rate $0.78$ and has the shortest mean time, path length, and highest mean speed among successful runs; and NeuPAN reaches success rate $0.76$. These results suggest that explicit footprint modeling can improve robustness in mixed static-dynamic clutter, even when competing methods achieve faster completion on their successful trials.

\begin{table}[t]
    \caption{Corridor dynamic-random benchmark (50 trials per method): success rate and mean metrics over successful runs for \algname{}, Convex-MPPI, Rectangle-MPPI, and NeuPAN with a convex-hull T-shaped footprint.}
    \label{tab:sim_anyshape_compare}
    \centering
    \setlength{\tabcolsep}{6pt}
    \renewcommand{\arraystretch}{1.05}
    \begin{adjustbox}{max width=\columnwidth}
        \begin{tabular}{lcccc}
            \toprule
            Method               & \shortstack{Success                                                    \\ rate} & \shortstack{Mean \\ Time (s)} & \shortstack{Mean \\ Path (m)} & \shortstack{Mean \\ Speed (m/s)} \\
            \midrule
            \algname{}           & \textbf{0.92}       & 44.25          & 63.62          & 1.482          \\
            Convex-MPPI          & 0.86                & 41.79          & 63.57          & 1.556          \\
            Rectangle-MPPI       & 0.78                & \textbf{39.66} & \textbf{63.00} & \textbf{1.609} \\
            NeuPAN (convex hull) & 0.76                & 42.95          & 64.05          & 1.491          \\
            \bottomrule
        \end{tabular}
    \end{adjustbox}
\end{table}

\subsubsection{Test Case 4: Gazebo Dynamic-Obstacle Comparison}

\begin{figure}[t]
    \centering
    \includegraphics[width=0.9\columnwidth]{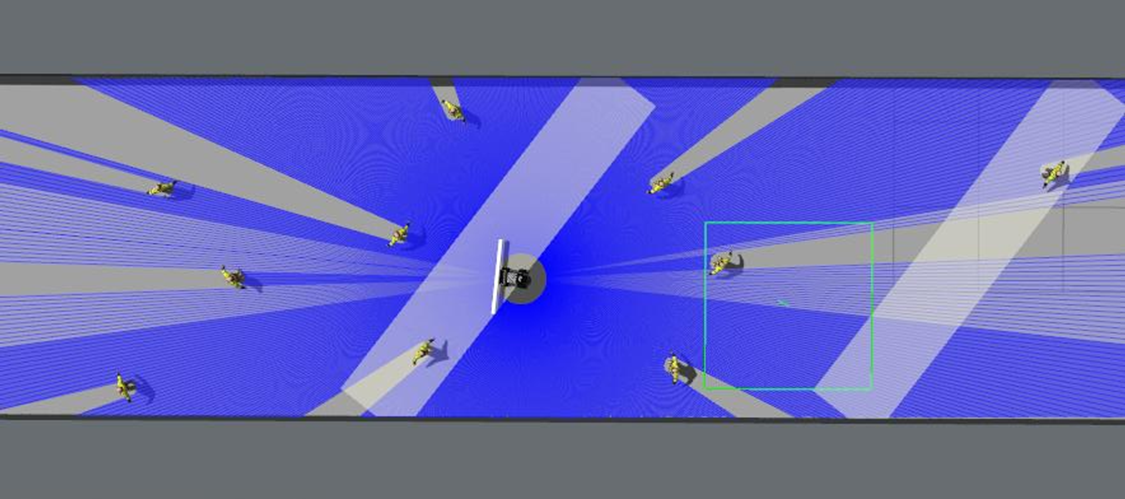}
    \caption{Gazebo dynamic-obstacle experiment setting.}
    \label{fig:gazebo_dynamic_setting}
\end{figure}

Following the Gazebo-style dynamic-obstacle setting used in the NeuPAN repository~\cite{neupan_ros_repo}, we construct a corridor benchmark with a differential-drive AgileX Limo carrying a 0.7~m extra load. The effective footprint used by the planner is T-shaped. The environment contains eight static obstacles randomly placed in the corridor and two moving obstacles, as shown in Fig.~\ref{fig:gazebo_dynamic_setting}. The moving obstacles follow fixed cross-corridor trails at random speeds below 0.2~m/s. For fairness, \algname{} and NeuPAN receive the same point-cloud observations and share the same vehicle-kinematics setting. NeuPAN is evaluated using the public implementation from its repository~\cite{neupan_ros_repo}.

For the loaded-Limo footprint, NeuPAN's DUNE distance encoder is trained using the convex hull of the effective T-shaped footprint. The NRMP planner parameters are initialized from the repository's tuned no-load dynamic-obstacle setting and then kept fixed for the added-load evaluation. This setup evaluates a practical add-on deployment scenario: after the robot footprint is changed by an extra load, the footprint-specific distance representation is updated, while the downstream planner configuration is transferred from the existing tuned setting.

We complete 50 Gazebo trials for each method. Table~\ref{tab:gazebo_dynamic_compare} summarizes the quantitative comparison between \algname{} and NeuPAN with a convex-hull T-shaped footprint. In this benchmark, \algname{} achieves a success rate of $0.96$, compared with $0.65$ for NeuPAN. Over successful runs, \algname{} also yields lower mean navigation time ($62.47$~s versus $64.47$~s) and shorter mean path length ($25.05$~m versus $26.49$~m), while NeuPAN has a slightly higher mean speed ($0.41$~m/s versus $0.40$~m/s). These results indicate that explicit footprint-aware signed-distance evaluation improves robustness in the setting with the added load while maintaining comparable efficiency on successful trials.

\begin{table}[t]
    \caption{Gazebo dynamic-obstacle benchmark (50 trials per method): success rate and mean metrics over successful runs for \algname{} and NeuPAN with a convex-hull T-shaped footprint.}
    \label{tab:gazebo_dynamic_compare}
    \centering
    \setlength{\tabcolsep}{6pt}
    \renewcommand{\arraystretch}{1.05}
    \begin{adjustbox}{max width=\columnwidth}
        \begin{tabular}{lcccc}
            \toprule
            Method               & \shortstack{Success                                                   \\ rate} & \shortstack{Mean \\ Time (s)} & \shortstack{Mean \\ Path (m)} & \shortstack{Mean \\ Speed (m/s)} \\
            \midrule
            \algname{}           & \textbf{0.96}       & \textbf{62.47} & \textbf{25.05} & 0.40          \\
            NeuPAN (convex hull) & 0.65                & 64.47          & 26.49          & \textbf{0.41} \\
            \bottomrule
        \end{tabular}
    \end{adjustbox}
\end{table}

\subsection{Experiment 3: Cross-Platform Deployment on Multiple Robot Platforms}

This experiment evaluates whether \algname{} can be deployed across heterogeneous robot platforms with limited platform-specific changes. The study is organized as three real-robot deployment case studies: an indoor differential-drive dual-arm transportation robot, the AgileX Ranger mini hybrid-motion platform, and a Unitree Go2 quadrupedal robot carrying an elongated object. In each case, the same collision-evaluation and MPPI update structure is retained, while the platform-specific motion model, effective-footprint representation, and sensor interface are updated. The goal of this experiment is to assess practical transferability of the framework, rather than to provide a statistically exhaustive benchmark for each platform.

The cross-platform motion-model and kinematic summary is given in Table~\ref{tab:platform_kinematic_setup}. The AgileX Ranger mini's dual-Ackermann, parallel, and spin modes are listed separately to make the hybrid-motion deployment explicit.

\subsubsection{Dual-Arm, Differential-Drive Transportation in Indoor Narrow Spaces}

We first deploy \algname{} on a dual-arm, differential-drive robot in an indoor transportation task. This case represents a conventional indoor service platform operating in a clearance-limited office environment. Front and rear 2D laser scanners provide local obstacle observations, and a 2D SLAM system~\cite{hess2016real} is used to transform the nominal guidance path into the robot coordinate frame.

\begin{figure}[t]
    \centering
    \includegraphics[width=1.0\columnwidth]{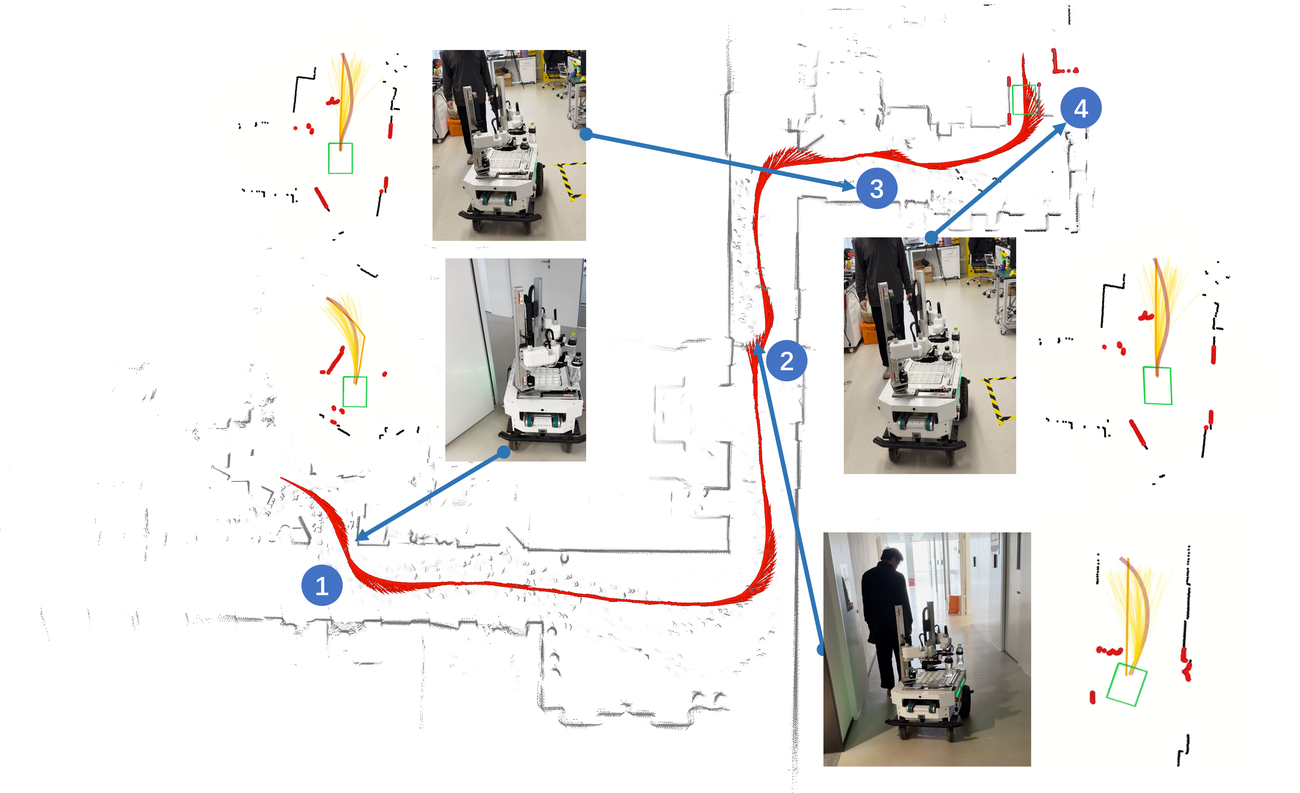}
    \caption{Differential-drive dual-arm transportation in an indoor office environment. The numbered sequence shows passage through a narrow gate, pedestrian interaction in a narrow corridor, and final entry into a tight workspace for bottle placement; the full task trajectory is also shown.}
    \label{fig:exper3_full}
\end{figure}

As shown in Fig.~\ref{fig:exper3_full}, the robot passes through a narrow gate, interacts with a pedestrian in a narrow corridor, and enters a tight workspace for bottle placement. The task is completed using the same point-cloud-based collision-evaluation and MPPI control structure as in simulation, with the rollout model adapted to differential-drive kinematics and the footprint specified for this platform. This deployment shows that the proposed collision-evaluation module can be reused on a conventional indoor service robot by updating the platform model and footprint description.

\subsubsection{AgileX Ranger mini Deployment in a Trap-Like Narrow-Space Scenario}

We next deploy \algname{} on the AgileX Ranger mini platform, which provides a complementary transfer case because it has a different footprint and supports multiple non-skidding motion modes. In this deployment, the footprint representation is replaced by the Ranger mini body geometry, and rollout propagation is matched to the available motion modes. The signed-distance evaluation and MPPI update structure remain unchanged. For the dual-Ackermann rollout model, we use half the axle-axis length as the effective wheelbase parameter.

\begin{figure}[t]
    \centering
    \includegraphics[width=0.8\columnwidth]{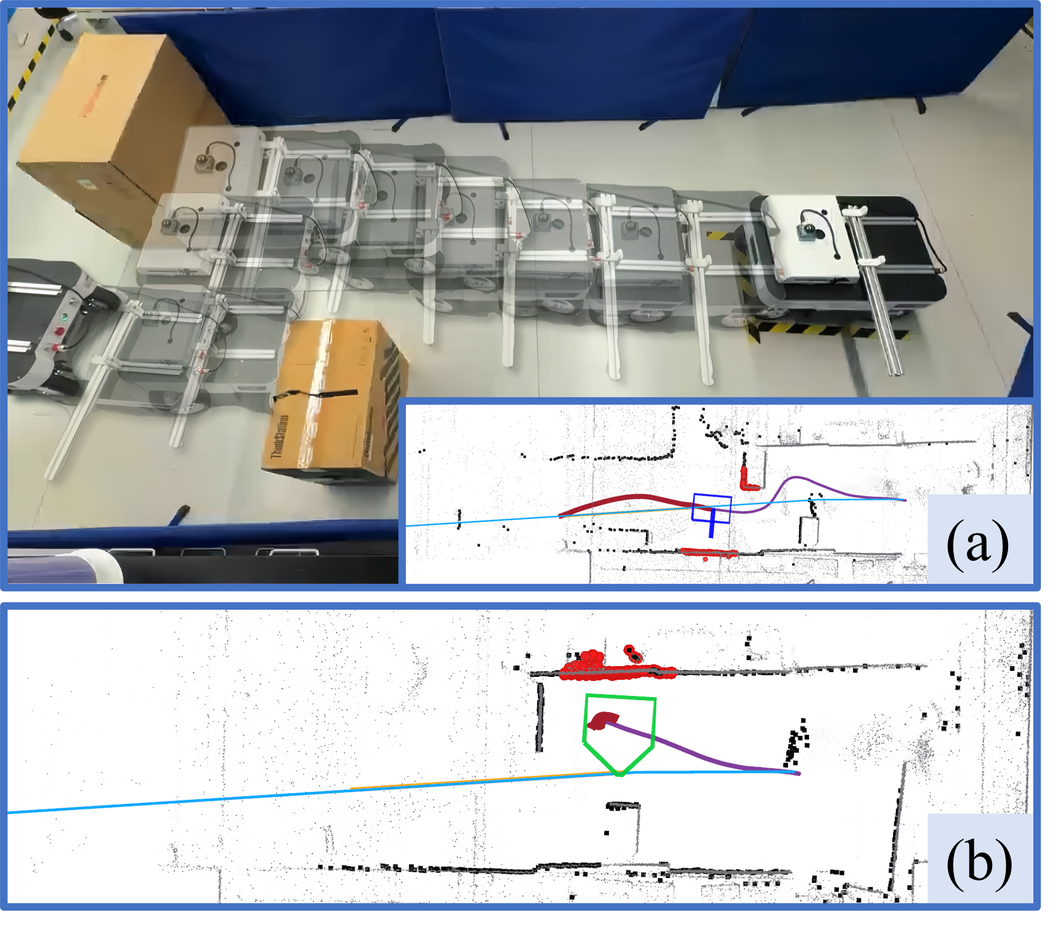}
    \caption{Ranger mini trap scenario under parallel motion: (a) \algname{} with the explicit footprint representation and (b) Convex-MPPI with the convex-hull footprint representation.}
    \label{fig:miniranger_parallel_exact_convex}
\end{figure}

Figure~\ref{fig:miniranger_parallel_exact_convex} compares \algname{} and Convex-MPPI in the parallel-motion case. With the explicit footprint representation, \algname{} exits the trap in approximately $9$~s. Convex-MPPI is unable to complete the same maneuver because the convex-hull approximation removes the narrow feasible passage at an effective degree of narrowness of $\mathrm{DoN}=1.0$. Here DoN is measured relative to the actual moving direction: under parallel motion, the robot translates sideways and cannot use rotation to reduce its effective cross section.

\begin{figure}[t]
    \centering
    \includegraphics[width=0.8\columnwidth]{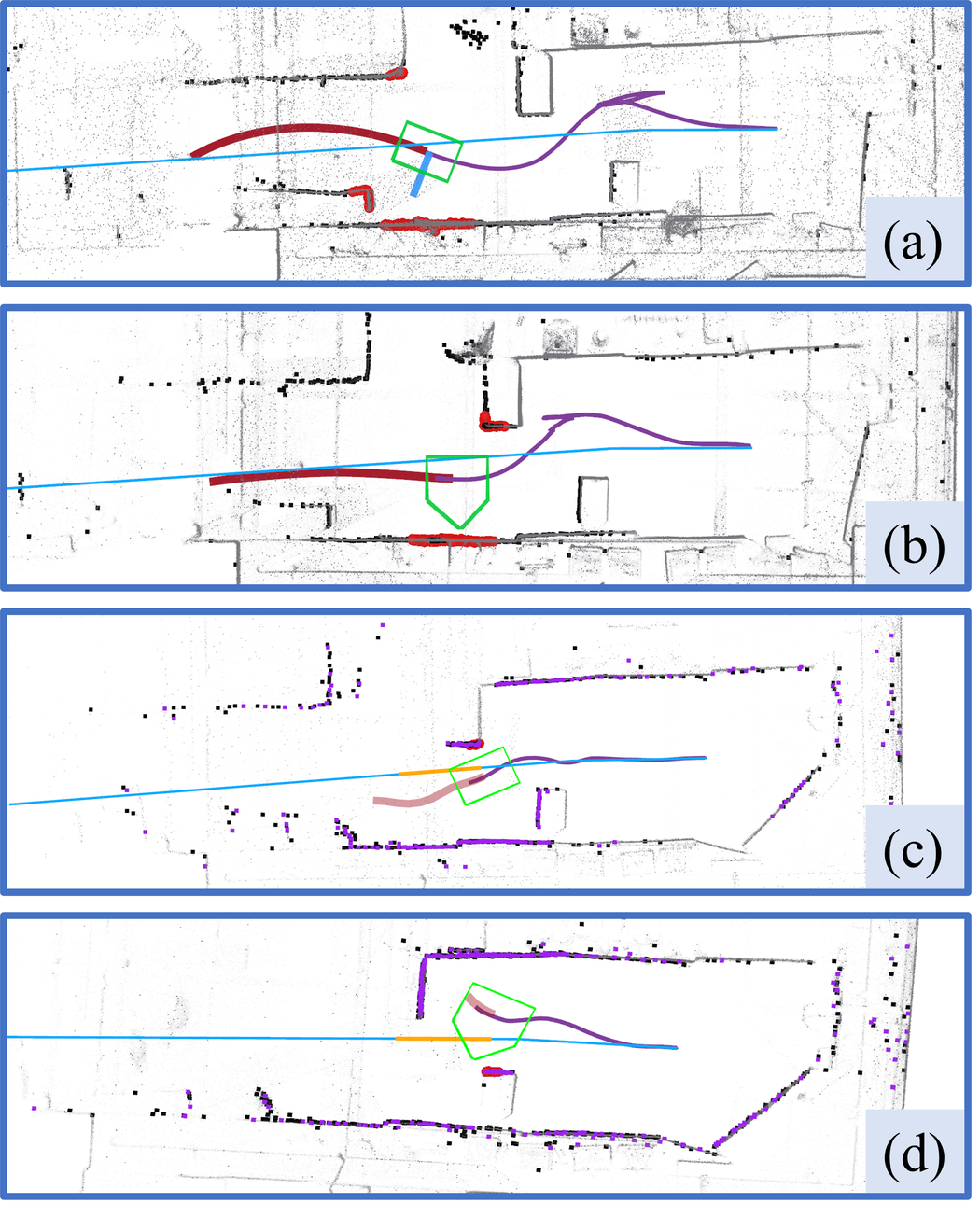}
    \caption{Ranger mini trap scenario under dual-Ackermann steering: (a) \algname{}, (b) Convex-MPPI, (c) NeuPAN with the nominal footprint configuration, and (d) NeuPAN with the added-load footprint configuration.}
    \label{fig:miniranger_ackermann_compare}
\end{figure}

Figure~\ref{fig:miniranger_ackermann_compare} compares the dual-Ackermann case across \algname{}, Convex-MPPI, and NeuPAN. Under dual-Ackermann steering, \algname{} exits the trap in approximately $13$~s, while Convex-MPPI requires approximately $35$~s in the same setting. In this real-robot comparison, \algname{} updates at approximately $30$~Hz, while NeuPAN runs at approximately $15$~Hz in our setup. Panels~(c) and~(d) show NeuPAN under two platform configurations. In Panel~(c), DUNE is trained for the nominal Ranger mini footprint and the NRMP planner parameters are tuned for the nominal chassis, allowing the robot to pass the trap. In Panel~(d), after attaching the extra load, DUNE is retrained for the updated footprint representation, while the NRMP parameters are transferred from the nominal-chassis setting. Under this transferred planner configuration, NeuPAN does not escape the trap. This comparison illustrates a practical deployment issue: when the task-dependent footprint changes, both the footprint-dependent distance representation and the downstream planner parameters may affect performance. In contrast, \algname{} updates the explicit footprint representation directly while preserving the same MPPI collision-evaluation structure.

\subsubsection{Unitree Go2 with a Carried Bar}

Finally, we deploy \algname{} on a Unitree Go2 quadrupedal robot carrying a rigid bar that extends the projected footprint beyond the nominal body width. This case tests transfer beyond wheeled robots to a legged platform that accepts body-velocity commands from the local planner. \algname{} uses the Go2 body-velocity interface together with a bar-augmented effective footprint, so the collision-evaluation module reasons about the carried object instead of only the nominal quadruped body.

\begin{figure}[t]
    \centering
    \includegraphics[width=0.75\columnwidth]{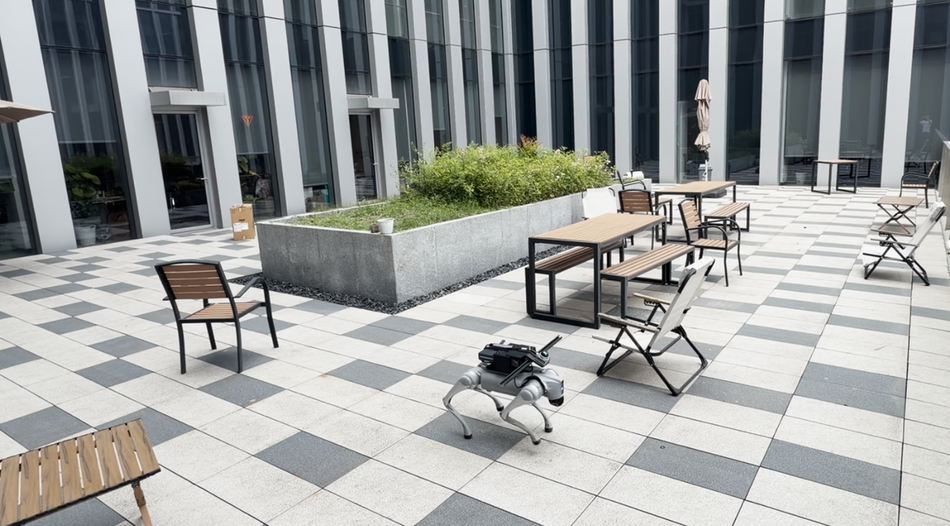}
    \caption{Garden test scenario for the Unitree Go2 carrying the extra bar in an unstructured outdoor environment.}
    \label{fig:go2_garden_test}
\end{figure}

We first test the full system in an outdoor garden-like environment with scattered tables, chairs, planters, and other irregular obstacles, as shown in Fig.~\ref{fig:go2_garden_test}. The robot follows a rectangular guidance path through the unstructured scene while carrying the extra bar. During execution, the local planner corrects the motion by evaluating the bar-augmented footprint against the observed point cloud. This case study evaluates end-to-end transportability of the perception-to-control loop on a legged body-velocity platform operating for the structured indoor layouts.

We also compare \algname{} with Falco~\cite{zhang2020falco} in the same garden scenario. Falco uses its standard rectangular robot abstraction, while \algname{} uses the bar-augmented footprint. \algname{} completes the traversal in $108.96$~s with a traveled distance of $45.72$~m, compared with $114.78$~s and $50.90$~m for Falco. During this comparison, \algname{} updates at approximately $30$~Hz, while Falco runs at approximately $50$~Hz. This comparison is intended to evaluate the effect of task-dependent footprint reasoning in this deployment case, rather than to isolate update frequency alone.

\begin{figure}[t]
    \centering
    \includegraphics[width=0.6\columnwidth]{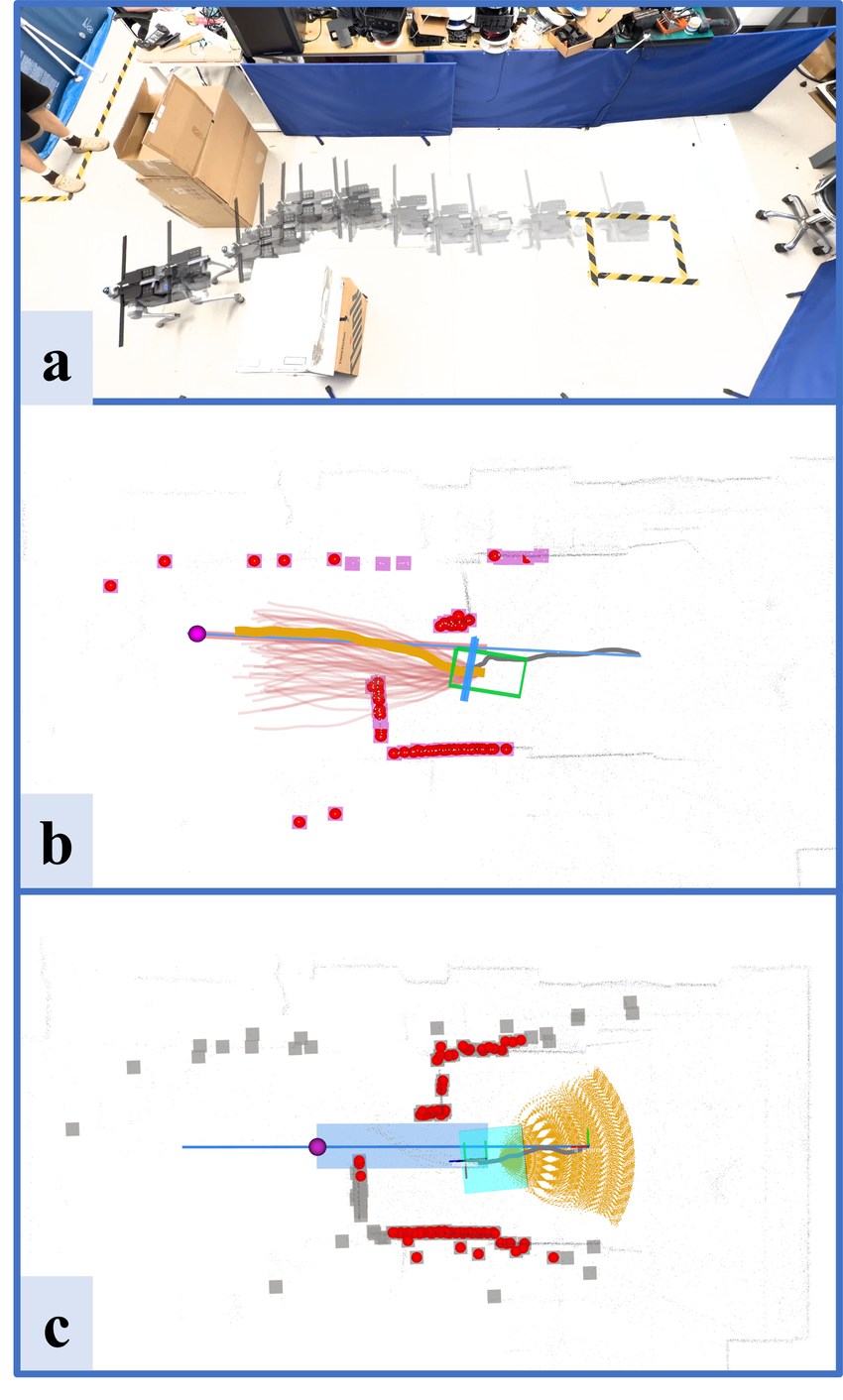}
    \caption{Extreme narrow-passage comparison for the Unitree Go2 with the carried bar: (a) elapsed traversal of \algname{}, (b) corresponding \algname{} planning result through the cluttered passage, and (c) Falco under the rectangular robot abstraction.}
    \label{fig:go2_narrow_case}
\end{figure}

We further construct an extreme narrow-passage case to isolate the effect of footprint modeling with the carried bar. As shown in Fig.~\ref{fig:go2_narrow_case}, \algname{} traverses the passage with the bar-augmented footprint, and the corresponding planning visualization shows a feasible path through the cluttered gap while maintaining explicit bar-aware clearance. Under the same scene, Falco with the rectangular robot abstraction does not complete the passage. This result is consistent with the broader observation that, in clearance-limited real-world navigation, explicit task-dependent footprint reasoning can preserve maneuvers that are lost under simplified footprint abstractions.

\begin{figure}[t]
    \centering
    \includegraphics[width=1.0\columnwidth]{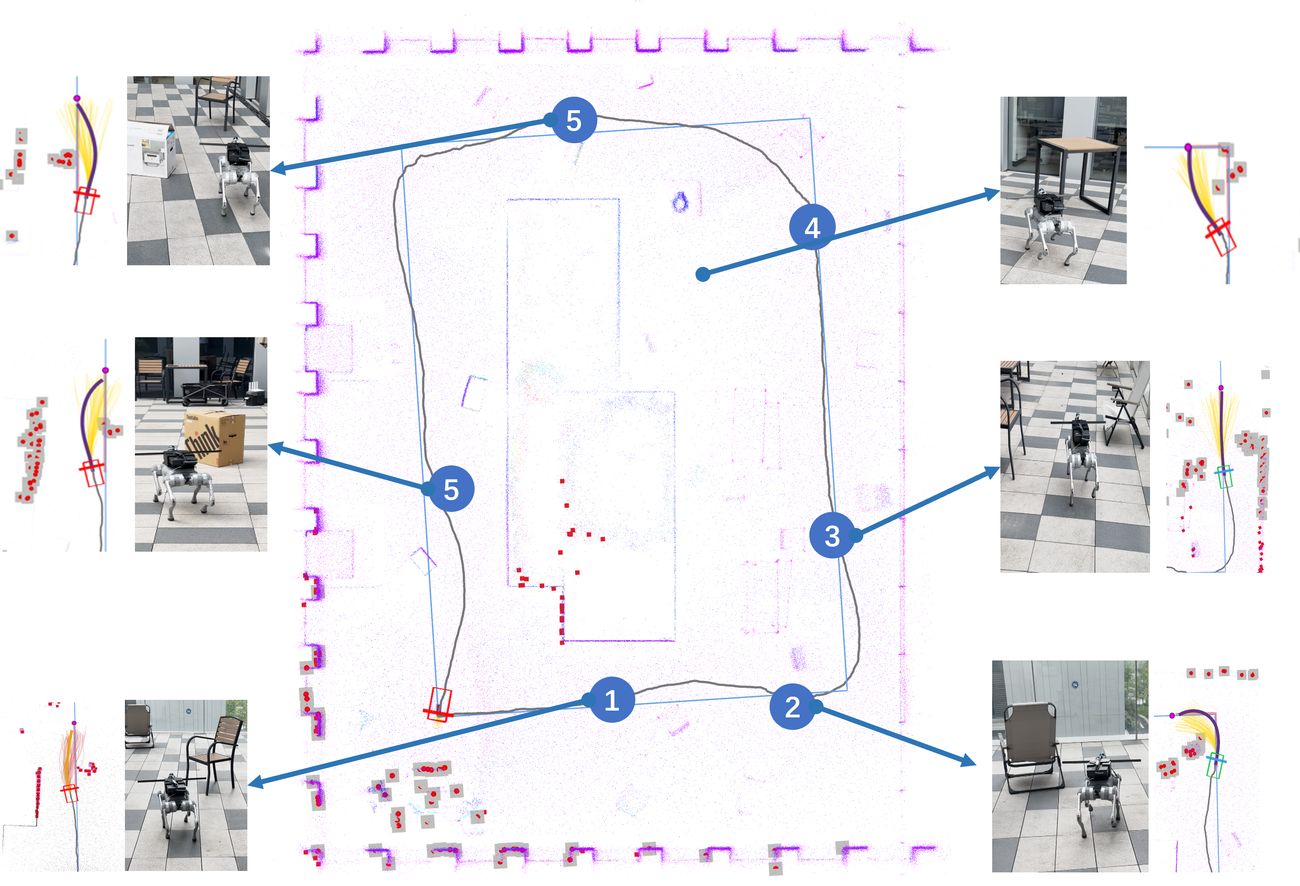}
    \caption{Unitree Go2 garden deployment with the carried bar: overall trajectory and representative local obstacle-avoidance snapshots during the run.}
    \label{fig:go2_garden_result}
\end{figure}

\begin{figure}[t]
    \centering
    \includegraphics[width=0.75\columnwidth]{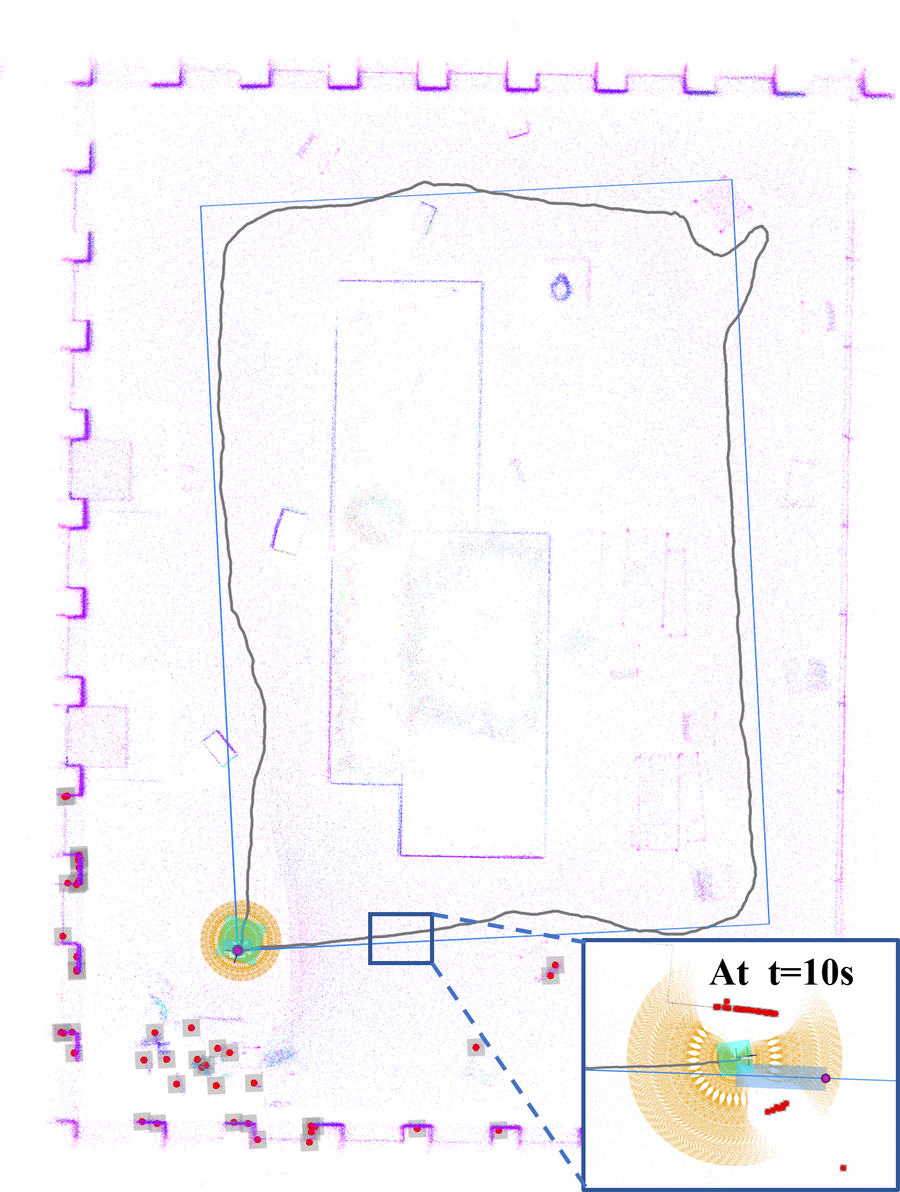}
    \caption{Falco in the same Unitree Go2 garden scenario with the carried bar.}
    \label{fig:go2_garden_falco}
\end{figure}

\subsection{Experiment 4: Hybrid-Motion Navigation on AgileX Ranger mini}

This experiment evaluates the hybrid-mode extension of \algname{} on the Ranger mini platform. While Experiment~3 demonstrated cross-platform deployment, this experiment isolates the contribution of multi-mode rollout selection by comparing the full hybrid-mode controller with a dual-Ackermann-only ablation under the same footprint model and environment.

\textbf{Experimental Setup.}
The test environment contains tight turns, narrow passages, and local recovery regions that require heterogeneous maneuvers. The representative narrow-space scenario has a maximum degree of narrowness of $\mathrm{DoN}=0.90$, so the task remains clearance-limited but is traversable with appropriate mode selection. The signed-distance evaluator, footprint representation, obstacle processing, and MPPI cost structure are kept the same for both methods. The only difference is the admissible motion set: the hybrid controller can select among dual-Ackermann, parallel, and spin-in-place modes, whereas the ablation is restricted to dual-Ackermann steering.

\textbf{Ablation Result.}
In the representative scenario, the dual-Ackermann-only configuration requires $140$~s to complete the task, while the hybrid-mode configuration completes it in $106$~s. This corresponds to an approximately $24\%$ reduction in completion time. The result indicates that access to multiple non-skidding motion modes can enlarge the local maneuver set and reduce unnecessary steering corrections in constrained regions.

\begin{figure*}[!t]
    \centering
    \includegraphics[width=0.85\textwidth]{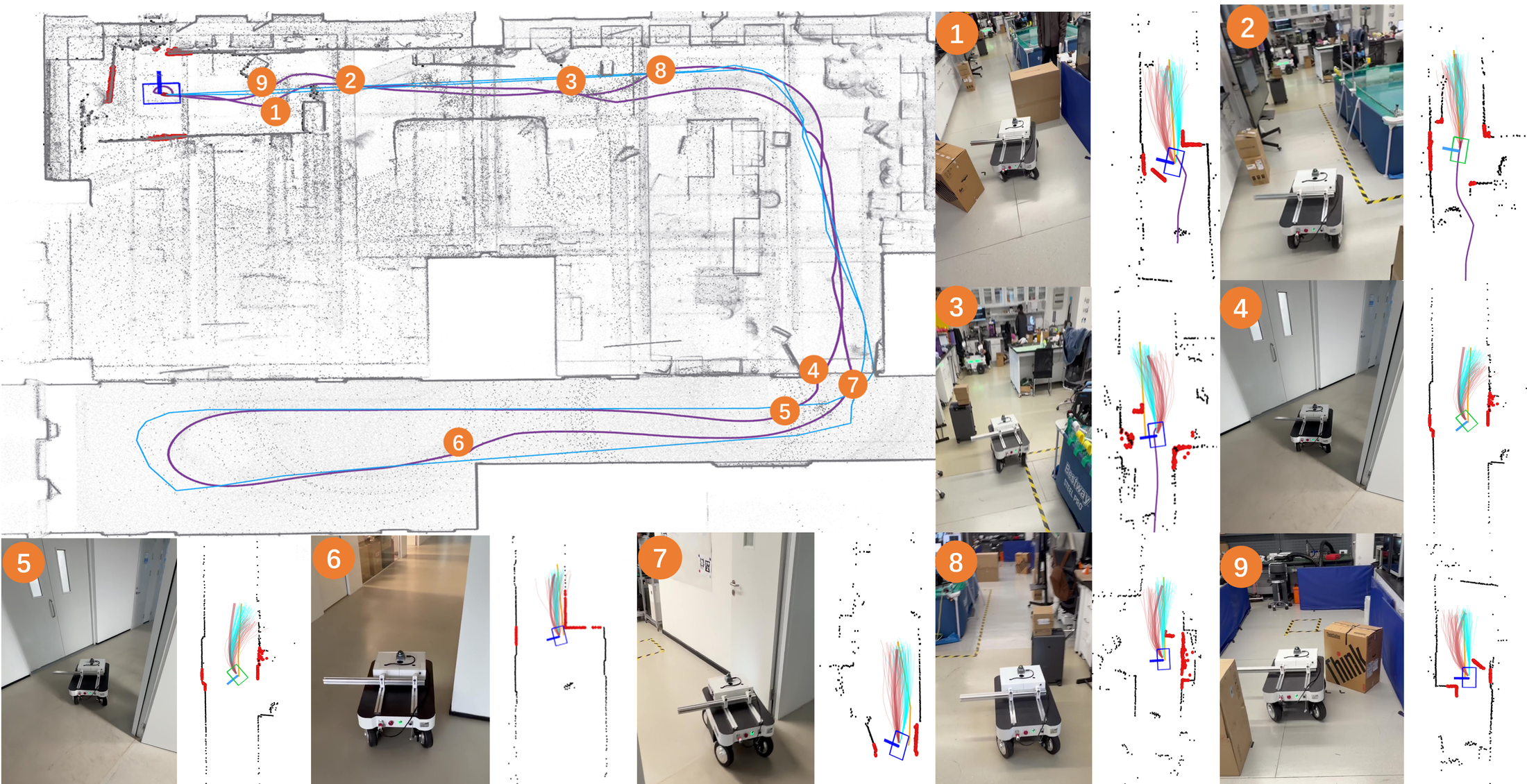}
    \caption{Hybrid-mode navigation on AgileX Ranger mini. The full trajectory is reconstructed for visualization with FAST-LIO2~\cite{xu2022fast}; local planning details show the point-cloud input, reference trajectory, sampled candidate paths, and selected MPPI trajectory at a representative replanning instant.}
    \label{fig:hybrid_mode_full_map}
\end{figure*}

\begin{figure*}[!t]
    \centering
    \includegraphics[width=0.85\textwidth]{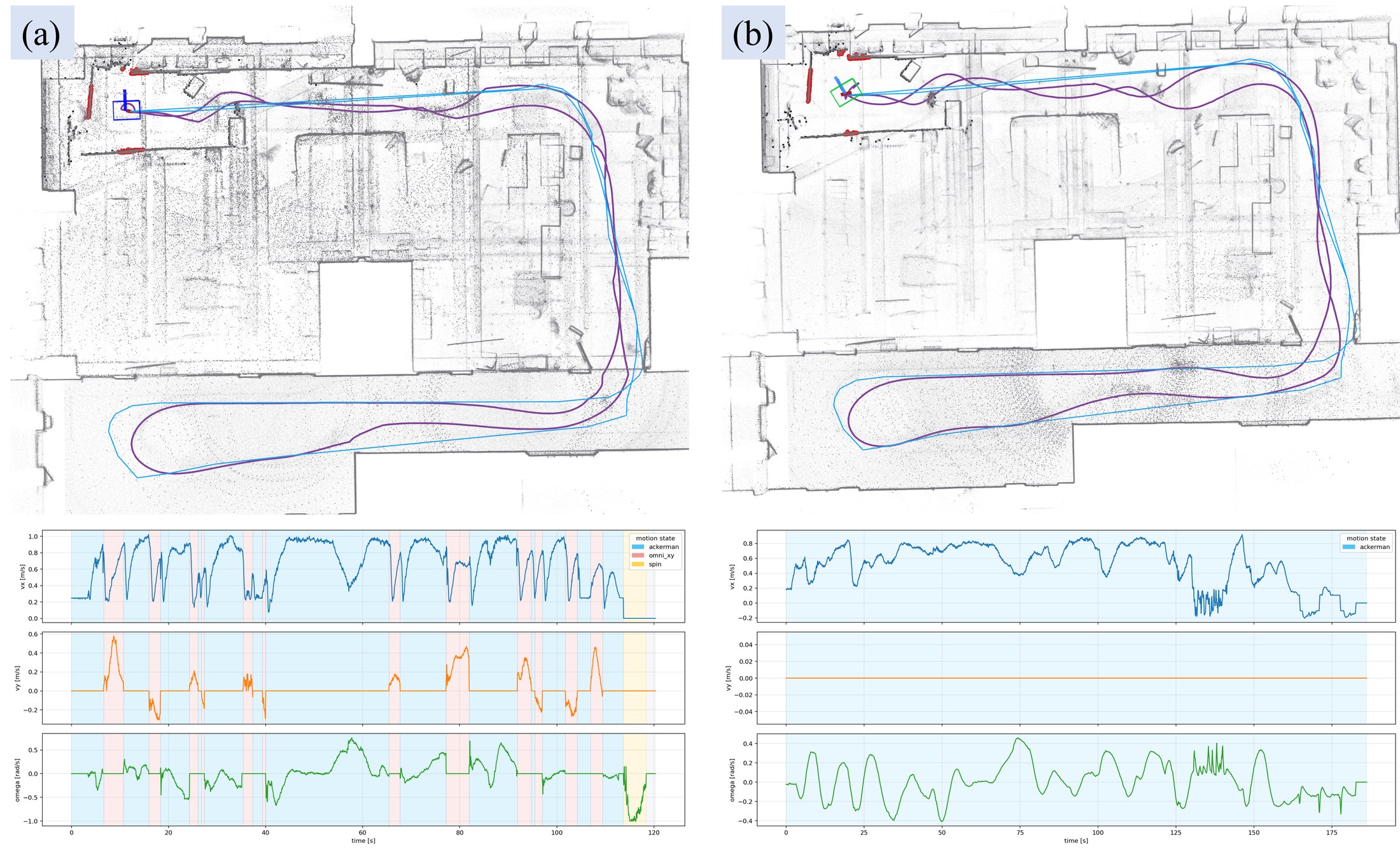}
    \caption{Trajectory and commanded velocities for (a) hybrid-mode operation and (b) dual-Ackermann-only operation.}
    \label{fig:trajplot_with_vcmd}
\end{figure*}

Figure~\ref{fig:hybrid_mode_full_map} visualizes the hybrid-mode traversal. The global trajectory is shown only for visualization; \algname{} does not use this reconstructed map for local planning. At each replanning step, the controller samples short-horizon candidates under the available motion modes, scores them with the footprint-aware signed-distance cost, and executes the first command of the selected sequence.

Figure~\ref{fig:trajplot_with_vcmd} compares hybrid-mode and dual-Ackermann-only operation. The velocity profiles show that the hybrid controller uses lateral velocity $v_y$ when parallel motion is advantageous and uses yaw-rate commands when rotation is needed. In contrast, the dual-Ackermann-only configuration must rely on forward motion and steering to negotiate the same constrained regions. This explains the shorter completion time of the hybrid-mode controller in this scenario.

\subsection{Discussion of Experimental Findings and Limitations}

The experiments support the central claim that explicit footprint-aware signed-distance evaluation can improve feasibility and robustness in clearance-limited settings. The narrow-passage studies show that convex-hull or simplified footprint models may remove feasible configurations when clearance is tight, even though they can be faster in some less restrictive successful trials. The dynamic-obstacle and hardware case studies indicate that the same collision-evaluation structure can be deployed across different platforms and sensing conditions. At the same time, the results should be interpreted within the scope of the assumptions used in this work.

First, \algname{} is a local planner that assumes obstacle observations and weak guidance are available from upstream modules. The planner does not address global route generation, semantic scene interpretation, or task-level decision making. In the experiments, the guidance signal is provided as a target pose or simple reference path, and the local planner is responsible for footprint-aware collision avoidance and short-horizon maneuver generation. Integrating the framework with richer perception modules, global planners, or learned navigation priors may improve behavior in more ambiguous scenes, but this system-level integration is outside the scope of the present study.

Second, rollout propagation is based on kinematic models. This choice is appropriate for the low-speed ground-navigation scenarios evaluated here, where the main challenge is local geometric clearance. However, the current validation does not establish dynamic feasibility for high-speed motion, aggressive maneuvers, rough-terrain locomotion, contact-rich legged locomotion, or articulated trailer-like systems. Extending the method to those settings would require replacing the kinematic rollout model with an appropriate dynamic or articulated model and adding stability, actuation, or articulation constraints while preserving the same footprint-aware collision-evaluation principle.

Third, collision evaluation is performed with respect to a planar projected footprint and a preprocessed local point-cloud observation. This is sufficient for the flat-ground scenarios considered in this paper, but it does not model full 3D body geometry, height-dependent clearance, overhanging obstacles, or posture-dependent robot shape. For robots operating in environments where vertical clearance or whole-body configuration matters, the current 2D footprint representation would need to be extended to 3D occupied-volume or configuration-dependent collision reasoning. Whether the same computational advantages carry over to such 3D formulations remains an open direction.

Finally, dynamic obstacles are handled through receding-horizon replanning from updated point-cloud observations, without explicit obstacle-motion prediction. This strategy is effective in the low-speed dynamic-obstacle scenarios tested in this work, but it may be insufficient in dense crowds or fast-changing environments where future obstacle motion must be anticipated. A natural extension is to incorporate time-indexed obstacle predictions into the rollout evaluation, so that the signed-distance cost is evaluated against predicted obstacle positions along the MPPI horizon rather than only the current observation.

\section{Conclusion}
\label{sec:conclusion}

This paper presented \algname{}, a perception-to-control local navigation framework that evaluates exact signed distances between observed point clouds and arbitrary 2D robot footprints, with motion commands generated directly by a sampling-based MPPI controller. The geometric evaluator handles both convex and concave footprints in a unified manner: orthogonal bodies are evaluated efficiently through a rectangle-cover point-to-box specialization, while general simple polygons are evaluated through analytic point-to-edge signed distance with an inside--outside test. The evaluator and the MPPI rollouts are fused into a single JAX-compiled GPU kernel, batched across rollout samples, horizon steps, obstacle points, and footprint edges, yielding a training-free and map-free local planning pipeline that supports real-time control.

The experiments support three conclusions within the scope considered in this work. First, the analytic signed-distance evaluator is well-matched to batched rollout evaluation on GPU and achieves an order-of-magnitude speedup over a learned point-to-robot distance baseline (Experiment~1). Second, explicit footprint-aware evaluation preserves feasible motion near clearance limits where convex approximations become overly conservative, while not necessarily producing the shortest completion time in every shared feasible setting. Third, the same evaluator transfers across multiple ground-robot platforms -- differential-drive, Ackermann-steering, omnidirectional, and hybrid -- without modification to the core geometric reasoning module. At the same time, the present results validate local navigation in planar settings with static or moderately dynamic obstacles, using kinematic rollout models and projected 2D footprints.

The contribution of this work is therefore best understood as a practical local-navigation framework rather than a complete solution covering perception, motion prediction, and whole-body planning. Future work will pursue tighter integration with higher-level perception and guidance modules, dynamic or articulated rollout models for more complex platforms (e.g., legged systems on rough terrain or articulated trailer-like vehicles), extension from planar footprints to 3D body-clearance reasoning, and explicit obstacle-motion prediction for more dynamic environments.


\bibliographystyle{IEEEtran}
\bibliography{IEEEabrv,reference}

\end{document}